\documentclass{article}

\usepackage{iclr2027_conference,times}

\iclrfinalcopy

\usepackage[utf8]{inputenc}
\usepackage[T1]{fontenc}
\usepackage{hyperref}
\usepackage{url}
\usepackage{booktabs}
\usepackage{amsfonts}
\usepackage{nicefrac}
\usepackage{microtype}
\usepackage{xcolor}
\usepackage{colortbl}
\definecolor{ourblue}{rgb}{0.368,0.507,0.71}
\hypersetup{colorlinks=true, linkcolor=ourblue, citecolor=ourblue, urlcolor=ourblue}

\usepackage{amsmath}
\usepackage{amssymb}
\usepackage{amsthm}
\newtheorem{proposition}{Proposition}
\usepackage{graphicx}
\usepackage{multirow}
\usepackage{subcaption}
\usepackage{enumitem}
\usepackage{wrapfig}
\usepackage{algorithm}
\usepackage{algorithmic}
\usepackage{tikz}
\usetikzlibrary{arrows.meta, positioning, calc}

\graphicspath{{figures/}}

\newcommand{\prism}{\textsc{Prism}}
\newcommand{\CR}{\mathrm{CR}}

\renewcommand{\arraystretch}{0.92}

\makeatletter
\def\section{\@startsection{section}{1}{\z@}{-1.2ex plus -0.4ex minus -.2ex}
  {0.8ex plus 0.2ex minus 0.1ex}{\large\sc\raggedright}}
\def\subsection{\@startsection{subsection}{2}{\z@}{-1.0ex plus -0.3ex minus -.2ex}
  {0.4ex plus .1ex}{\normalsize\sc\raggedright}}
\def\paragraph{\@startsection{paragraph}{4}{\z@}{0.9ex plus 0.3ex minus .2ex}
  {-1em}{\normalsize\bf}}
\makeatother

\title{Predicting and Repairing Merge Collapse\\ in Large Language Models\vspace{14pt}}

\author{%
  \vspace*{4pt}
  \makebox[5.28in][c]{%
  \begin{tabular}{@{}c@{\hspace{0.02in}}c@{\hspace{0.02in}}c@{}}
    \parbox[c]{1.76in}{\centering Jungseob Lee$^{1}$\\[1pt] \fontsize{8pt}{7.5pt}\texttt{omanma1928@korea.ac.kr}} &\vspace*{4pt}
    \parbox[c]{1.76in}{\centering Seungyoon Lee$^{1}$\\[1pt] \fontsize{8pt}{7.5pt}\texttt{dltmddbs100@korea.ac.kr}} &\vspace*{4pt}
    \parbox[c]{1.76in}{\centering Sugyeong Eo$^{2}$\\[1pt]
    \fontsize{8pt}{7.5pt}\texttt{s.eo@yonsei.ac.kr}}\\[4pt]\vspace*{4pt}
    \parbox[c]{1.76in}{\centering Hyeonseok Moon$^{3}$\\[1pt] \fontsize{8pt}{7.5pt}\texttt{hyns.moon@sookmyung.ac.kr}} &
    \parbox[c]{1.76in}{\centering Jaehyung Seo$^{4,\dagger}$\\[1pt] \fontsize{8pt}{7.5pt}\texttt{seojae777@konkuk.ac.kr}} &
    \parbox[c]{1.76in}{\centering Heuiseok Lim$^{1,\dagger}$\\[1pt] \fontsize{8pt}{7.5pt}\texttt{limhseok@korea.ac.kr}}
  \end{tabular}%
  }\\[7pt]
  \small
  \makebox[5.28in][c]{%
    \begin{tabular}{c}
      $^{1}$Korea University \hspace{0.1in} $^{2}$Yonsei University Mirae Campus\\[1pt]
      $^{3}$Sookmyung Women's University \hspace{0.1in} $^{4}$Konkuk University
    \end{tabular}%
  }%
}

\begin{document}

\maketitle
\begingroup
\renewcommand{\thefootnote}{\textdagger}
\footnotetext{Corresponding authors.}
\endgroup

\lhead{Preprint}

\begin{abstract}Large language models fine-tuned from a shared base can be merged by averaging their task vectors, but some merges collapse far below the base model, and common merge operators give no warning before evaluation. We show that one statistic of the specialists' task vectors both predicts this collapse and calibrates its repair. The power that averaging removes equals the variance of the task vectors across specialists, our measure of interference. Under a working noise model, the disturbance that a merge injects grows with the merge coefficient and with interference, yielding a pre-merge score. In our experiments on twenty-two merge configurations from four model families, only destructive merges exceed a threshold on this score. We find that statistics of sign conflict between specialists, a common target of existing merge operators, are anti-predictive. We then predicted the outcomes of fourteen merges before evaluating them, and twelve predictions were correct, including the destructive outcome of a specialist pair pushed past the threshold by continued pretraining. To address this collapse, we introduce \prism{}, an operator that averages the task vectors first and then soft-thresholds each layer at a level set by the layer's interference. Without data or tuning, \prism{} keeps all five destructive merges above the threshold within evaluation noise of the base model, where plain averaging falls at least 14.4 points below it or collapses entirely. We apply \prism{} only above the threshold and keep the plain average for merges below it, which include all fifteen harmless ones. Code is available at \url{https://github.com/js-lee-AI/PRISM}.\end{abstract}

\section{Introduction}
\label{sec:intro}

Model merging combines several models fine-tuned from the same pretrained checkpoint into one model by operating directly on their weights, with no further training \citep{wortsman2022model,matena2022merging,ilharco2022patching}. Task Arithmetic \citep{ilharco2022editing} represents each fine-tuned specialist by its task vector, the difference between its weights and those of the shared base model. A merge adds a scaled combination of task vectors back to the base, in the simplest case their average multiplied by a merge coefficient \citep{ilharco2022editing,ortizjimenez2023tangent}. Open-weight large language models (LLMs) now arrive as families of specialists for mathematics, code and other domains, built on shared bases \citep{yang2024qwen25,grattafiori2024llama,yang2024qwen2,hui2024qwen2,cao2026qwen3coder}. Merging such specialists has become common practice, including in the training of production models \citep{cohere2025command}, and is supported by dedicated toolkits, benchmarks and studies at scale \citep{goddard2024arcee,tang2025mergebench,yadav2024matters,wang2026scaling,hitit2026systematic}.

When specialists are fine-tuned for different tasks, however, they change shared weights in different directions, and averaging cancels part of each change, an effect usually called interference \citep{yadav2023ties,gargiulo2025task}. We call a merge destructive when it falls substantially below the base model from which its specialists were fine-tuned, and harmless when it stays near or above that base. The standard way to reduce interference is to discard part of each task vector before averaging \citep{yadav2023ties,yu2024language,deep2024della}. Each such prune-then-merge operator sets its keep rate in advance, whatever the disagreement between the specialists. These operators, like methods that tune the merge on data \citep{matena2022merging,yang2023adamerging,chopra2025lewis} or derive it from the task vectors alone \citep{marczak2025iso,cheng2025whoever}, leave open whether a merge will be destructive until it has been built and evaluated.

We start from the identity underlying the ambiguity decomposition of ensembles \citep{krogh1994neural}. At every weight coordinate, the mean-squared power that averaging removes from the task vectors equals their variance across specialists. This variance is our measure of interference, and we compute it from the weights in the same pass as the average. Interference splits into cancellation, where specialists write comparable values of opposite sign, and dispersion, the remainder. Dispersion includes one-sided drift, in which one specialist moves a weight that the others leave near zero, and such drift can hold knowledge unique to that specialist \citep{wang2024localizing,lu2024twin}. Under a working noise model, we show that the disturbance a merge injects into the base grows in proportion to the merge coefficient and to the square root of the interference. We call this estimate of the disturbance, available before merging, the interference score.

We introduce \prism{}, a merge operator whose denoising threshold is set by interference, as Figure~\ref{fig:teaser} illustrates. The operator averages the task vectors first and then soft-thresholds the average in each layer at the universal threshold of wavelet shrinkage \citep{donoho1994ideal,donoho1995noising}. The noise level comes from the layer's interference, a floor keeps a small share of each matrix's coordinates, and one threshold scale serves all settings without data or tuning. A gate on the share of interference that comes from cancellation returns the plain average when that share is small, leaving one-sided drift intact.

Across twenty-two merge configurations from four model families, with up to eight specialists in one merge, each outcome is either destructive or harmless. Seven merges are destructive at the default coefficient, six of them between public specialists. A threshold on the interference score flags five of the seven and none of the fifteen harmless merges, whereas statistics of sign disagreement, the conflict that sign election resolves \citep{yadav2023ties}, are anti-predictive. We then fixed the score and this threshold, predicted the outcomes of fourteen further merges before evaluation, and found twelve predictions correct.

Specialists are often built by continuing to pretrain a base model on domain text \citep{gururangan2020dont,hui2024qwen2}, and a practitioner who plans to merge them needs to know when more training would make the merge destructive. We therefore built two such specialists from one base model, one on Japanese web text and the other on biomedical abstracts, and tracked the pair's score as training proceeded. At a fixed learning rate, the score grew roughly as the square root of the number of training tokens. Continued training at a higher learning rate further improved each specialist on its own domain but pushed the pair past the threshold. Plain averaging then became destructive as predicted, while \prism{} stayed near the base.

\prism{} keeps each destructive merge above the threshold within evaluation noise of the base, where plain averaging falls far below it. The merge we study most closely combines a math specialist with a code specialist, as recent merging work does \citep{yu2024language,cheng2025whoever,tang2025mergebench}. On this merge, \prism{} leads Task Arithmetic, two prune-then-merge operators \citep{yadav2023ties,deep2024della} and a method that tunes the merge on data \citep{chopra2025lewis} at a shared coefficient. When we sweep the merge coefficient, these baselines collapse at large values while \prism{} degrades gradually. A merge whose coefficients are tuned by entropy minimization on unlabeled text \citep{yang2023adamerging} also falls far below the base at the shared coefficient. Shrinkage helps a destructive merge but can forgo the gain that plain averaging brings to a harmless one, and our recipe therefore averages plainly below the score threshold and applies \prism{} above it. The score needs only the weights and minutes of CPU time, and we hope it makes a pre-merge check routine whenever public checkpoints are combined.

\section{Related Work}
\label{sec:related}
Task Arithmetic \citep{ilharco2022patching,ilharco2022editing} established linear task-vector combination, and subsequent work has refined merging along three lines \citep{liu2026globa,cheng2025whoever,wang2024localizing,gargiulo2025task}. Prune-then-merge methods include TIES-Merging, which combines magnitude trimming with cross-model sign election \citep{yadav2023ties}, DARE, which drops entries at random and rescales the survivors \citep{yu2024language}, and DELLA, which drops entries with a probability that decreases with their magnitude \citep{deep2024della}.

Training-free approaches include model matching \citep{xu2024training}, hierarchical merging \citep{fu2025training}, and trust-region constraints \citep{sun2025task}. Fisher averaging \citep{matena2022merging}, AdaMerging \citep{yang2023adamerging}, LEWIS \citep{chopra2025lewis} and RAIN-Merging \citep{huang2026rain} use auxiliary data. NeuroMerging decomposes task representations into neuronal subspaces \citep{fang2025see}, and NeuronMerge groups neurons by function \citep{gu2025neuronmerge}. EMR-Merging \citep{huang2024emr} uses task-specific masks, which requires knowing the task at test time.

Spectral and subspace methods use spectrum flattening \citep{marczak2025iso}, task singular vectors \citep{gargiulo2025task}, or task-vector guidance for data-free interference reduction \citep{cheng2025whoever}. PACT preprocesses task vectors through subspace alignment and offers a randomized-SVD variant \citep{park2026pact}. Twin-Merging dynamically combines shared and exclusive knowledge according to the input \citep{lu2024twin}.

We screen for collapse risk, a step absent from the merge operators evaluated here. The \prism{} repair soft-thresholds the coordinates of the averaged task vector in linear time, rather than pruning each task vector or computing singular value decompositions. It sets the threshold from the weights alone, through an exact variance identity and a noise model.
\section{Interference in Task-Vector Merging}
\label{sec:theory}

\begin{figure}[t]
\centering
\includegraphics[width=0.9\linewidth]{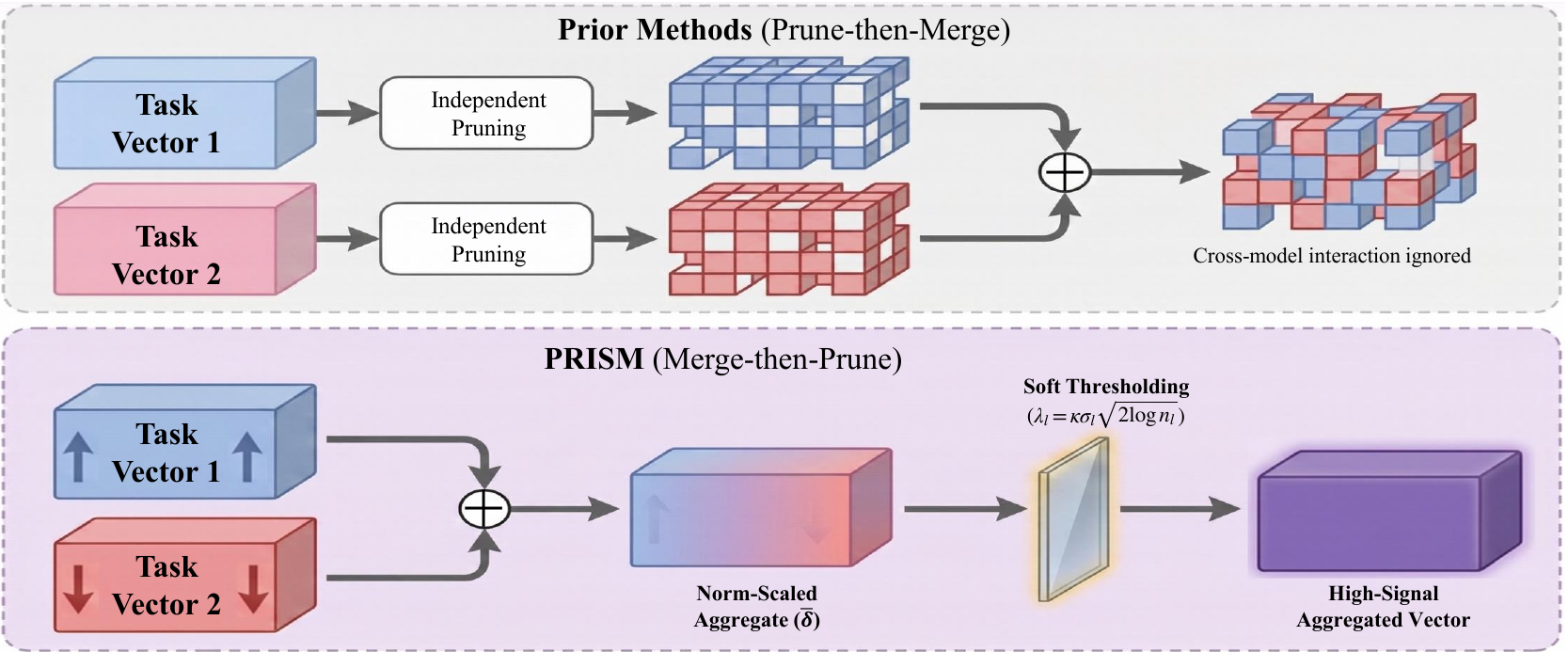}
\caption{Merge then prune. \textbf{Top:} task vectors are pruned before aggregation. \textbf{Bottom:} \prism{} averages first with norm scaling and then soft-thresholds the aggregate at the layer's interference scale. Full algorithm in Appendix~\ref{app:algorithm}.}
\label{fig:teaser}
\end{figure}

\subsection{Interference is the cross-model variance}
\label{sec:identity}

Let $\theta_{\text{base}}$ denote the parameters of a pretrained base model and $\theta_k$ those of specialist $k\in\{1,\dots,K\}$, fine-tuned from it. The task vector $\tau_k = \theta_k - \theta_{\text{base}}$ has value $\tau_k(i)$ at coordinate $i$. A merge rescales each task vector by a weight $s_k>0$, set by norm scaling in Section~\ref{sec:scope} or to one in a raw merge, and averages them into $\bar{\delta} = \frac{1}{K}\sum_k s_k \tau_k$. At coordinate $i$ the total power $P(i) = \frac{1}{K}\sum_k (s_k \tau_k(i))^2$ splits into signal power $S(i) = \bar{\delta}(i)^2$ and \textit{interference} $I(i) = P(i) - S(i)$, the reduction in mean-squared power due to averaging. 
\begin{equation}
I(i) = \underbrace{\tfrac{1}{K}\sum_{k=1}^{K}\!\big(s_k \tau_k(i)\big)^2}_{P(i)} \;-\; \underbrace{\bar{\delta}(i)^2}_{S(i)} \;=\; \text{Var}_k\!\big(s_k \tau_k(i)\big) \geq 0 ,
\label{eq:interference_var}
\end{equation}
where $\text{Var}_k$ is the population variance over the $K$ specialists. Equation~\ref{eq:interference_var} is the ambiguity decomposition of \citet{krogh1994neural} applied to the scaled task vectors, with the base model as the target. This variance measures task-vector disagreement, which we use as a proxy for task interference \citep{yadav2023ties,gargiulo2025task}, and is computed alongside the mean. Interpreting it as noise requires a model. Throughout, \emph{drift} means the displacement $s_k\tau_k$ of a specialist from the base. Identical scaled task vectors give zero interference regardless of the size or effect of their shared drift. Aligned vectors with unequal magnitudes can still have positive variance, as can \emph{one-sided} drift, present in one specialist and absent in the others.

Its normalized read-out is the \textit{Coherence Ratio}, $\CR(i) = S(i) / (P(i) + \epsilon) \in [0, 1]$ with $\epsilon = 10^{-10}$, a continuous generalization of sign agreement that stays meaningful when task vectors differ in magnitude. For a layer $l$ with $n_l$ merged coordinates, the means $\overline{P}_l$, $\overline{S}_l$ and $\overline{I}_l$ of $P$, $S$ and $I$ over those coordinates give $\overline{\CR}_l = \overline{S}_l / (\overline{P}_l + \epsilon)$ and an interference scale
\begin{equation}
\sigma_l^2 = \overline{I}_l = \overline{P}_l - \overline{S}_l \simeq \overline{P}_l(1 - \overline{\CR}_l) .
\label{eq:sigma_est}
\end{equation}
The last expression omits the stabilizer term $-\epsilon\overline{\CR}_l$. We compute $\sigma_l$ from $\overline{P}_l-\overline{S}_l$. The standard deviation $\sigma_l=\sqrt{\overline{I}_l}$ has weight units and can set a threshold on task-vector coordinates. Computing these statistics takes about two CPU-minutes on a 7B model. Sign agreement tracks $\overline{\CR}_l$ almost exactly at the layer level ($r\!=\!0.998$) but is a dimensionless fraction, so the threshold uses $\sigma_l$ to supply the weight scale, as shown in Appendix~\ref{app:cr_vs_sa}.

\subsection{What a merge injects}
\label{sec:injected}

The merge applies $c\,\bar\delta$, where $c$ is an additional global multiplier on the already-averaged task vector. Our default $c=1/K$ further shrinks that mean, while $c=1$ applies it in full.

\begin{proposition}[Injected interference]
\label{prop:amplify}
Assume that, at a coordinate of layer $l$, the scaled task vectors share a signal and have independent, zero-mean disturbances of common variance $v_l$. For a nonnegative coefficient $c$, adding $c\,\bar\delta$ introduces disturbance standard deviation $c\sqrt{v_l/K}$ and power $c^2v_l/K$. Using $\overline{I}_l$ as an empirical proxy for $v_l$ gives the screening scale $c\sqrt{\overline{I}_l/K}$ and power $c^2\overline{I}_l/K$, both available before merging.
\end{proposition}

\noindent\textit{Argument, with the full version in Appendix~\ref{app:amplify_proof}.} Independence removes cross-covariance terms, giving disturbance variance $v_l/K$ for the mean. Scaling by $c$ multiplies this variance by $c^2$. Substituting $\overline{I}_l$ for $v_l$ is our modeling choice. \hfill$\square$

The mean update is an absolute perturbation of the base weights. Its estimated disturbance power grows as $c^2$ at fixed interference, the regime that the coefficient sweep of Section~\ref{sec:c_sweep} examines.

\subsection{Cancellation and dispersion}
\label{sec:cancel}

Treating all interference as noise may suppress useful specialist knowledge \citep{wang2024localizing,lu2024twin}. Low coherence has two causes. \textit{Cancellation} is specialists writing comparable magnitudes with opposing signs so the aggregate nearly vanishes. \textit{Dispersion} includes one specialist writing a large value while the others sit near zero, task-exclusive drift that can carry useful signal. With $\text{Pos}(i)=\sum_k \max(s_k\tau_k(i),0)$ and $\text{Neg}(i)=\sum_k \max(-s_k\tau_k(i),0)$ the positive and negative mass at coordinate $i$, the cancellation power is
\begin{equation}
C(i) = \min\!\Big(\,\tfrac{2}{K}\min\big(\text{Pos}(i),\text{Neg}(i)\big)^2,\; I(i)\Big), \qquad D(i) = I(i) - C(i) \ge 0,
\label{eq:cancel_disp}
\end{equation}
so that $I = C + D$. Summing over all merged coordinates yields the \emph{empirical cancellation ratio} $\rho = \big(\sum_l n_l\overline{C}_l\big)/\big(\sum_l n_l\overline{I}_l+\epsilon_g\big)\in[0,1]$, where $\overline{C}_l$ is the mean of $C$ over layer $l$ and $\epsilon_g=10^{-12}$ is a separate gate stabilizer. High $\rho$ marks a conflict-dominated merge and low $\rho$ a dispersion-dominated one, and empirically $\rho$ spans $0.003$--$0.720$ across four families. The interference scale sets the initial threshold, and $\rho$ determines whether shrinkage is applied.
\section{Repairing a Dangerous Merge with \prism{}}
\label{sec:method}

\prism{} averages first and shrinks the aggregate using the interference scale $\sigma_l$ of Eq.~\ref{eq:sigma_est}, as in Figure~\ref{fig:teaser}.

\subsection{Interference-calibrated soft thresholding}
\label{sec:denoise}

Unlike prune-then-merge methods \citep{yadav2023ties,deep2024della}, \prism{} averages first and then denoises the consensus with the Donoho--Johnstone universal soft threshold \citep{donoho1994ideal}, used here as a tuning-free default,
\begin{equation}
\tilde{\delta}(i) = \operatorname{sgn}(\bar{\delta}(i)) \max(0,\, |\bar{\delta}(i)| - \lambda_l), \quad \text{where} \quad \lambda_l = \kappa \sigma_l \sqrt{2 \log n_l}.
\label{eq:soft_thresh}
\end{equation}
Substituting Eq.~\ref{eq:sigma_est} gives $\lambda_l = \kappa \sqrt{2 \overline{I}_l \log n_l}$ with the threshold scale $\kappa$ fixed at $1.0$. At fixed total power, high-coherence layers ($\overline{\CR}_l \to 1$) have smaller interference scales and thresholds. A coordinate survives only when its aggregate magnitude exceeds $\lambda_l$, and its magnitude is then reduced by $\lambda_l$ in Algorithm~\ref{alg:prism}. \prism{} shrinks the aggregate once and then scales the resulting vector by $c$. When fewer than $0.1\%$ of a projection matrix's coordinates survive, a keep-rate floor lowers the threshold to the smallest magnitude among its top $0.1\%$ and applies soft thresholding again.

\vspace{-2pt}

\subsection{The \texorpdfstring{$\rho$}{rho}-gate}
\label{sec:gate}

Calibrating by $\sigma_l=\sqrt{\overline{I}_l}$ treats interference as destructive conflict. For one-sided drift, where one specialist moves while the others do not, $\sigma_l$ can be large even when the update carries useful specialist signal. The cancellation ratio $\rho$ of Section~\ref{sec:cancel} measures how much interference comes from opposing signs, so we gate on it with a threshold $\tau_g$,
\begin{equation}
\lambda_l = \begin{cases} \kappa\,\sigma_l\sqrt{2\log n_l} & \text{if } \rho \ge \tau_g, \\[2pt] 0 & \text{if } \rho < \tau_g, \end{cases}
\qquad \tau_g = 0.1.
\label{eq:rho_gate}
\end{equation}
Under substantial cancellation ($\rho\ge\tau_g$), \prism{} applies soft thresholding. Under dispersion ($\rho<\tau_g$), it sets $\lambda_l=0$, leaves the mean unchanged, and yields the Task-Arithmetic model with the same scaling and coefficient. Fallback examples from two families include the raw Llama-3.1-8B $K\!=\!3$ merge at $\rho\!=\!0.021$, the Mistral-7B $K\!=\!2$ merge at $\rho\!=\!0.093$ and Tulu-3 with its own direct preference optimization (DPO) variant at $\rho\!=\!0.003$. With the reported ratios, any threshold strictly between $0.093$ and $0.229$ makes the same decisions as the default $\tau_g\!=\!0.1$ on every setting here.

\subsection{Merge procedure and scope}
\label{sec:scope}

The merged model is $\theta_{\text{merged}} = \theta_{\text{base}} + c \tilde{\delta}$ with norm scaling $s_k = \min_j \|\tau_j\|_2 / \|\tau_k\|_2$, where the norms run over the merged coordinates. We apply \prism{} to the multilayer perceptron (MLP) projections (\texttt{gate/up/down\_proj}) and leave attention at its base values in the default comparison, since MLP layers are the primary locus of domain-specific knowledge, holding key--value memories with localized knowledge neurons~\citep{geva2022transformer,dai2022knowledge}, where interference should concentrate. Restricting \prism{} to the MLP projections stays within $0.1$pp of the six-task mean obtained by extending it to attention, which merges $2.3\times$ more keys, as reported in Appendix~\ref{app:implementation}.
\section{Experiments}
\label{sec:experiments}

\subsection{Setup}
\label{sec:setup}

As in MergeBench \citep{tang2025mergebench}, we vary model family and specialist domain, and we also vary the number of merged specialists. We merge specialists from two 7--8B families, Qwen2.5 \citep{yang2024qwen25} and Llama-3.1 \citep{grattafiori2024llama}, adding Mistral-7B and DeepSeek-7B for the breadth study. The main setting pairs Qwen2.5 Math \citep{yang2024qwen2} with Coder \citep{hui2024qwen2} at $K\!=\!2$, extended to $K\!=\!3$ (Math, Coder, Instruct). For the shared task-vector protocol, we norm-scale by $s_k = \min_j \|\tau_j\|_2 / \|\tau_k\|_2$ before averaging. By default, we apply an additional global coefficient $c = 1/K$ to that average under Appendix~\ref{app:amplify_proof}'s convention.

Baselines are Task Arithmetic \citep{ilharco2022editing}, TIES \citep{yadav2023ties}, DELLA \citep{deep2024della}, DARE \citep{yu2024language}, LEWIS \citep{chopra2025lewis} with the activation-based variant and weight-sparsity fallback compared in Appendix~\ref{app:lewis_ctp}, AdaMerging \citep{yang2023adamerging}, Greedy Soup \citep{wortsman2022model}, Fisher-Weighted Averaging \citep{matena2022merging} and the 2025 data-free Iso-C \citep{marczak2025iso} and WUDI-Merging \citep{cheng2025whoever}. Headline task-vector comparisons use the shared coefficient, with baseline settings and sweeps summarized in Appendix~\ref{app:fairness}, while \prism{} stays at its data-free default $\kappa\!=\!1.0$. SLERP \citep{shoemake1985animating} and Model Stock \citep{jang2024modelstock} use their published recipes without our coefficient or norm scaling.

Our evaluation spans six benchmarks, GSM8K \citep{cobbe2021training}, ARC-Challenge \citep{clark2018think}, MMLU \citep{hendrycks2021measuring}, TruthfulQA \citep{lin2022truthfulqa}, HellaSwag \citep{zellers2019hellaswag} and WinoGrande \citep{sakaguchi2020winogrande}, under \texttt{lm-evaluation-harness} \citep{gao2024language}, plus WikiText-2 perplexity \citep{merity2016pointer}, with GSM8K under strict-match extraction. Every reported mean is the average of the six task scores shown, and differences are absolute percentage points (pp). Generation metrics on GSM8K, MATH-500 \citep{lightman2023lets}, HumanEval \citep{chen2021evaluating} and MBPP \citep{austin2021program} use greedy decoding with one sample for each problem. A six-task mean carries a propagated standard error of at most $0.8$pp, so we treat sub-pp gaps as ties under Appendix~\ref{app:multi_seed}'s uncertainty estimate.

\subsection{Merging under severe conflict}
\label{sec:k2_results}

In Table~\ref{tab:k2_results}, \prism{} leads Task Arithmetic, TIES and DELLA on the six-task mean by at least $6.7$pp and exceeds calibration-guided LEWIS without using calibration data. Task Arithmetic falls to $13.4$ on HumanEval, whereas \prism{} leads every baseline on all four generation metrics, by at least $25.0$pp on HumanEval. With instruction-tuned specialists, \prism{} again leads these baselines in Appendix~\ref{app:multi_seed} and posts the best training-free coding scores in Appendix~\ref{app:coding}.

\paragraph{Model scale.} Table~\ref{tab:1.5b_main} repeats the setting on Qwen2.5-1.5B. Task Arithmetic collapses outright, and \prism{} leads the strongest baseline by $3.5$pp. Appendix~\ref{app:scale} adds Instruct$+$Coder-Instruct merges at 3B and 14B, where the plain averages that the recipe keeps gain $0.8$ and $1.0$pp over the base.

\paragraph{Vendor-native coding evaluation.} Under vendor-native \textsc{EvalPlus} evaluation, Task Arithmetic, TIES, DELLA, DARE, LEWIS and SLERP score at most $0.3$ on all four coding metrics in Table~\ref{tab:extended_baselines}, while Model Stock reaches $0.6$. \prism{} retains coding capability across these metrics. Section~\ref{sec:controls} compares \prism{} with the base and a norm-matched control on these metrics.

\begin{table}[t]
\caption{Qwen2.5-7B Math$+$Coder base checkpoints. Generation: GSM8K 5-shot, MATH-500 4-shot, harness HumanEval/MBPP pass@1 with completion prompts. LEWIS uses activation-based six-task/HumanEval/MBPP results and weight-sparsity GSM8K/MATH generation. Bold: best merged; underline: second best; $^\dagger$: calibration for activation-based LEWIS.}
\label{tab:k2_results}
\centering
\setlength{\tabcolsep}{2.6pt}
\footnotesize
\resizebox{\linewidth}{!}{
\begin{tabular}{l cccccc c cccc}
\toprule
& \multicolumn{6}{c}{\textbf{Six-task evaluation}} & & \multicolumn{4}{c}{\textbf{Generation}} \\
\cmidrule(lr){2-7} \cmidrule(lr){9-12}
\textbf{Method} & GSM8K & ARC & MMLU & TQA & HellaS. & WinoG. & \textbf{Avg.} & GSM8K & MATH & HumanE. & MBPP \\
\midrule
\rowcolor{gray!8} Math specialist & 84.4 & 49.2 & 60.2 & 47.9 & 58.8 & 68.4 & 61.5 & 83.8 & 53.6 & 34.1 & 56.2 \\
\rowcolor{gray!8} Coder specialist & 81.8 & 49.0 & 69.5 & 49.8 & 66.4 & 72.0 & 64.8 & 82.4 & 41.4 & 59.8 & 68.4 \\
\midrule
Task Arithmetic & 22.0 & 47.0 & 55.7 & 50.0 & 55.4 & 66.6 & 49.5 & 23.0 & \phantom{0}5.8 & 13.4 & 27.4 \\
TIES & \underline{67.2} & \textbf{53.4} & 64.3 & 50.2 & 60.6 & \underline{71.0} & \underline{61.1} & 65.8 & \underline{24.6} & 17.7 & 43.0 \\
DELLA & 65.2 & \underline{51.6} & 64.2 & 50.3 & 59.8 & 70.4 & 60.3 & 64.0 & 22.0 & 17.7 & 42.4 \\
LEWIS$^\dagger$ & 43.0 & 44.2 & \underline{69.1} & \underline{50.7} & \underline{62.8} & 66.6 & 56.1 & \underline{74.2} & 15.4 & \underline{26.2} & \underline{52.6} \\
\rowcolor{gray!15} \textbf{\prism{} (ours)} & \textbf{81.4} & 51.4 & \textbf{74.3} & \textbf{55.2} & \textbf{67.4} & \textbf{77.0} & \textbf{67.8} & \textbf{82.4} & \textbf{30.2} & \textbf{51.2} & \textbf{62.8} \\
\bottomrule
\end{tabular}}
\end{table}

\begin{table}[t]
\caption{Qwen2.5-1.5B Math$+$Coder instruction-tuned checkpoints. Columns, marks and LEWIS variants follow Table~\ref{tab:k2_results}.}
\label{tab:1.5b_main}
\centering
\setlength{\tabcolsep}{2.6pt}
\footnotesize
\resizebox{\linewidth}{!}{
\begin{tabular}{l cccccc c cccc}
\toprule
& \multicolumn{6}{c}{\textbf{Six-task evaluation}} & & \multicolumn{4}{c}{\textbf{Generation}} \\
\cmidrule(lr){2-7} \cmidrule(lr){9-12}
\textbf{Method} & GSM8K & ARC & MMLU & TQA & HellaS. & WinoG. & \textbf{Avg.} & GSM8K & MATH & HumanE. & MBPP \\
\midrule
Task Arithmetic & \phantom{0}4.6 & 32.4 & 35.7 & \textbf{49.4} & 43.4 & 57.0 & 37.1 & \phantom{0}4.0 & \phantom{0}1.8 & \phantom{0}0.0 & \phantom{0}0.0 \\
TIES & 18.8 & 37.6 & 41.2 & 46.1 & 50.6 & 61.2 & 42.6 & 19.0 & \phantom{0}3.8 & 11.6 & 16.2 \\
LEWIS$^\dagger$ & \underline{49.4} & \underline{40.2} & \underline{57.8} & \underline{46.4} & \underline{55.4} & \underline{64.8} & \underline{52.3} & \underline{35.6} & \underline{13.6} & \underline{28.0} & \underline{38.6} \\
\rowcolor{gray!15} \textbf{\prism{} (ours)} & \textbf{59.6} & \textbf{42.6} & \textbf{61.8} & 45.5 & \textbf{59.0} & \textbf{66.4} & \textbf{55.8} & \textbf{59.0} & \textbf{26.6} & \textbf{35.4} & \textbf{43.0} \\
\bottomrule
\end{tabular}}
\end{table}

\subsection{Robustness to the scaling coefficient}
\label{sec:c_sweep}

\begin{figure}[t]
\centering
\begin{minipage}[t]{0.485\linewidth}
\centering
\includegraphics[width=\linewidth]{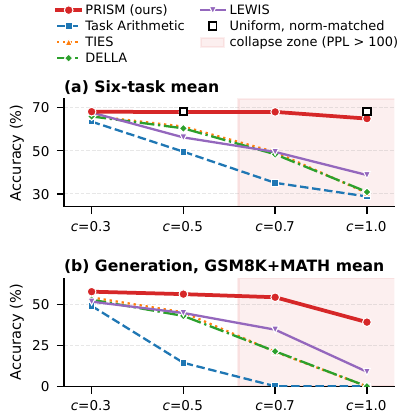}
\caption{Coefficient sweep on Qwen2.5-7B Math$+$Coder: \textbf{(a)} six-task evaluation and \textbf{(b)} separately prompted generation. Squares mark the norm-matched control.}
\label{fig:c_sweep}
\end{minipage}\hfill
\begin{minipage}[t]{0.485\linewidth}
\centering
\includegraphics[width=\linewidth]{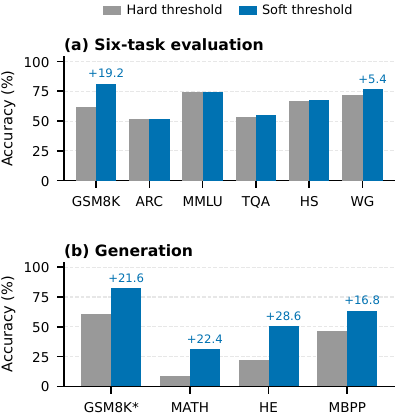}
\caption{Soft and hard thresholding at the same interference scale on Qwen2.5-7B Math$+$Coder. Labels mark gaps above five points; GSM8K$^{*}$ uses the generation-format evaluation.}
\label{fig:soft_hard}
\end{minipage}
\end{figure}

Proposition~\ref{prop:amplify} motivates a $c^2$ increase in injected interference power under its working model. \prism{} shrinks the update before applying the coefficient. In Figure~\ref{fig:c_sweep}, \prism{} holds a $64.8$--$68.0\%$ mean across the sweep. At $c\!=\!0.7$, the four evaluated baselines sit $19$ to $33$ points below the base at perplexities from $100$ to $961$. In the generation evaluation at $c\!=\!1.0$, Task Arithmetic, TIES and DELLA score zero on GSM8K, while LEWIS reaches $17.4$ in Appendix~\ref{app:csweep_harness}. The best baseline in the plotted sweep, LEWIS at $67.6$ ($c\!=\!0.3$), is tied within evaluation uncertainty with the $67.8$ that untuned \prism{} posts at the shared default.

\begin{table}[t]
\caption{Qwen2.5-7B Math$+$Coder. Coding: vendor-native \textsc{EvalPlus} pass@1 at $c\!=\!0.5$ where applicable, not comparable with Table~\ref{tab:k2_results}. LEWIS: activation-based Mean/PPL, weight-sparsity fallback for Norm/Coords/Coding. Bold: above the unmerged base. $^{\S}$: rescaled; $^{\dagger}$: calibration for activation-based LEWIS; SLERP and Model Stock: no coefficient.}
\label{tab:extended_baselines}
\centering
\scriptsize
\renewcommand{\arraystretch}{1.0}
\setlength{\tabcolsep}{5.02pt}
\begin{tabular}{lccccccccc}
\toprule
& \textbf{Norm} & \textbf{Coords} & \multicolumn{2}{c}{\textbf{Mean}} & \textbf{PPL} & \multicolumn{4}{c}{\textbf{Coding (EvalPlus)}} \\
\cmidrule(lr){4-5}\cmidrule(lr){7-10}
\textbf{Method} & $\|\delta\|/\|\bar\delta\|$ & \% & $c\!=\!0.5$ & $c\!=\!1$ & $c\!=\!1$ & HumanEval & HumanEval+ & MBPP & MBPP+ \\
\midrule
\rowcolor{gray!8} Base model, unmerged & 0 & 0.0 & 68.0 & 68.0 & 8.6 & 62.8 & 57.9 & 69.8 & 59.5 \\
\midrule
Task Arithmetic & 1.00 & 99.7 & 49.5 & 28.8 & $4.6\!\times\!10^{5}$ & 0.0 & 0.0 & 0.0 & 0.0 \\
TIES & 0.80 & 34.8 & 61.1 & 30.4 & $1.1\!\times\!10^{4}$ & 0.0 & 0.0 & 0.0 & 0.0 \\
DELLA & 0.82 & 34.9 & 60.3 & 30.9 & $1.6\!\times\!10^{4}$ & 0.0 & 0.0 & 0.0 & 0.0 \\
DARE (drop $0.9$)$^{\S}$ & 2.85 & 19.0 & 30.0 & 29.8 & $7.2\!\times\!10^{6}$ & 0.0 & 0.0 & 0.0 & 0.0 \\
LEWIS$^{\dagger}$ & 0.55 & 5.0 & 56.1 & 38.7 & $5.9\!\times\!10^{3}$ & 0.0 & 0.0 & 0.3 & 0.0 \\
SLERP ($t{=}0.5$) & 1.27 & 99.9 & \multicolumn{2}{c}{30.0} & $1.0\!\times\!10^{6}$ & 0.0 & 0.0 & 0.0 & 0.0 \\
Model Stock & 0.46 & 99.6 & \multicolumn{2}{c}{53.2} & 35.0 & 0.6 & 0.6 & 0.0 & 0.0 \\
\midrule
Uniform, norm-matched & 0.022 & 85.1 & 68.1 & 68.0 & 8.7 & 62.2 & 57.3 & 69.6 & 59.0 \\
Hard threshold, same $\lambda$ & 0.126 & 0.1 & 63.4 & 57.2 & 18.4 & 51.8 & 48.2 & 60.1 & 51.6 \\
\rowcolor{gray!15} \prism{} & 0.022 & 0.1 & 67.8 & 64.8 & 11.1 & \textbf{65.9} & \textbf{59.1} & \textbf{71.2} & \textbf{61.1} \\
\bottomrule
\end{tabular}
\end{table}

\subsection{Is it just conservative shrinkage?}
\label{sec:controls}

We test whether generic shrinkage \citep{yu2024language} explains the preserved capability with a control matched to \prism{}'s update norm. Table~\ref{tab:extended_baselines} reports each update's norm relative to the plain average. The baseline operators and community defaults write updates with $46$ to $285\%$ of that norm, while \prism{} writes $2.2\%$. For the matched control, one scalar multiplies every coordinate of the plain average to reach the same norm.

The norm-matched control avoids collapse at both evaluated coefficients and stays close to the unmerged base on the six-task mean, perplexity and all four coding metrics. Its norm, however, is taken from \prism{}, whose thresholds set that norm from each layer's interference without data or tuning. Where the control spreads this norm across $85.1\%$ of the coordinates, \prism{} concentrates it on the $0.1\%$ retained by thresholding. \prism{}'s coding scores are higher than both the control's and the base's on all four \textsc{EvalPlus} metrics, reaching $65.9$ on HumanEval against $62.2$ and $62.8$.

\subsection{Data-dependent baselines need data to reach parity}
\label{sec:extended}

AdaMerging reaches $67.9$ at its best point, a tie with \prism{}'s $67.8$, after entropy minimization on unlabeled C4 text and a coefficient sweep. Greedy Soup adds neither specialist and returns the base model, and Fisher-weighted averaging falls to $47.5$ in Appendix~\ref{app:fisher_soup}. Among data-free competitors, Iso-C reaches $66.6$ when tuned away from its collapsing default, and WUDI-Merging diverges in every configuration of Appendix~\ref{app:why_baselines}. SLERP at $t\!=\!0.5$ \citep{shoemake1985animating,goddard2024arcee} writes $1.27\times$ the norm-scaled average and collapses. Model Stock \citep{jang2024modelstock} applies $2.4\times$ interpolation shrinkage toward the base and still loses $14.8$pp.

\subsection{Ablations and mechanism}
\label{sec:ablation}
\label{sec:analysis}

On Qwen $K\!=\!2$, soft thresholding beats hard truncation at the same threshold by $4.4$pp on the six-task mean in Table~\ref{tab:extended_baselines}. Truncation's cost falls almost entirely on tasks that produce text, at $19.2$pp on GSM8K and up to $28.6$pp on HumanEval against $1.4$pp averaged over the five multiple-choice benchmarks in Figure~\ref{fig:soft_hard}. Across threshold scales from $\kappa\!=\!0.25$ to $3$, the six-task mean varies by at most $0.5$pp, and the default scores highest in Table~\ref{tab:ablation}. Replacing soft thresholding with magnitude pruning of the average at coherence-adapted layer rates lowers the mean by $3.7$pp at 7B and $6.9$pp at 1.5B, and Appendix~\ref{app:prune_comparison} shows that the gap closes at $K\!=\!3$, where interference is lower.

Appendix~\ref{app:mechanism} examines the layer-retention pattern. Where the calibrated threshold rather than the floor sets the keep rate ($\kappa\!=\!0.5$), layer retention $p_l$ tracks coherence data-free at $r\!=\!0.91$. Low coherence mixes sign conflict with task-exclusive drift, which the split of Eq.~\ref{eq:cancel_disp} separates, and the $\rho$-gate returns the plain average when dispersion dominates. Appendix~\ref{app:rescue} shows that the coordinates removed by thresholding are predominantly conflict elements. In our low-rank simulation using SVD projections of full task vectors, \prism{} stays within $0.6$pp of its full-weight six-task mean across $r\!\in\!\{8,\dots,128\}$ in Appendix~\ref{app:lora}.

\subsection{The same scale supports pre-merge screening}

\begin{figure}[t]
\centering
\includegraphics[width=\linewidth]{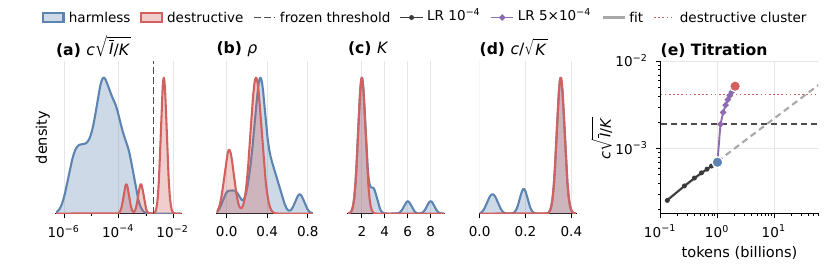}
\caption{\textbf{(a)} Densities of the interference score by outcome. \textbf{(b--d)} Densities of $\rho$, $K$ and $c/\sqrt{K}$ overlap. \textbf{(e)} At the lower learning rate, the score grows roughly as the square root of specialist training tokens. Large markers are evaluated merges.}
\label{fig:predictor}\label{fig:titration}
\end{figure}
\label{sec:k3_k1}
\label{sec:breadth}

The same $\overline{I}$ that sets the threshold defines the pre-merge score $c\sqrt{\overline{I}/K}$ motivated by Proposition~\ref{prop:amplify}. We test whether this score separates destructive and harmless merges.

\begin{table}[t]
\caption{The interference score exceeds the threshold of $1.9\!\times\!10^{-3}$ only for destructive merges. Means are six-task accuracies, and self-SFT specialists are fine-tuned by us from the base. $^{\star}$Outcome predicted before evaluation. $^{g}$Gated, so \prism{} equals Task Arithmetic (TA).}
\label{tab:breadth}
\centering
\scriptsize
\renewcommand{\arraystretch}{0.72}
\setlength{\tabcolsep}{6.56pt}
\begin{tabular}{lccccccc}
\toprule
\textbf{Setting} & $K$ & $c$ & $\rho$ & $c\sqrt{\overline{I}/K}$ $(\times10^{-3})$ & \textbf{Base} & \textbf{TA} & \textbf{\prism{}} \\
\midrule
\multicolumn{8}{l}{\textit{naive averaging destroys the model}} \\
Qwen2.5-1.5B ja\,$+$\,med, constructed$^{\star}$ & 2 & 0.50 & 0.342 & 5.19 & 56.1 & 35.8 & 55.9 \\
Qwen2.5-1.5B Math\,$+$\,Coder & 2 & 0.50 & 0.246 & 4.94 & 56.3 & 37.1 & 55.8 \\
Qwen2.5-7B Math\,$+$\,Coder & 2 & 0.50 & 0.293 & 4.25 & 68.0 & 49.5 & 67.8 \\
Qwen2.5-7B R1-Distill\,$+$\,Coder$^{\star}$ & 2 & 0.50 & 0.291 & 4.24 & 68.1 & 49.7 & 67.7 \\
Qwen2.5-7B Math-PRM\,$+$\,Coder$^{\star}$ & 2 & 0.50 & 0.285 & 4.17 & 68.1 & 53.7 & 67.6 \\
Qwen2.5-7B Math\,$+$\,Finance$^{\star}$ & 2 & 0.50 & 0.036$^{g}$ & 0.653 & 68.1 & 37.0 & 37.0$^{g}$ \\
Qwen2.5-7B Coder\,$+$\,Coder-Instruct$^{\star}$ & 2 & 0.50 & 0.013$^{g}$ & 0.190 & 68.1 & 39.3 & 39.3$^{g}$ \\
\midrule
\multicolumn{8}{l}{\textit{merging is harmless}} \\
DeepSeek-7B math\,$+$\,chat & 2 & 0.50 & 0.348 & 0.228 & 46.8 & 48.1 & 46.6 \\
Qwen2.5-7B self-SFT at $10\times$ learning rate$^{\star}$ & 2 & 0.50 & 0.349 & 0.227 & 68.1 & 68.8 & 68.2 \\
Llama-3.1-8B R1-Distill\,$+$\,Tulu-3$^{\star}$ & 2 & 0.50 & 0.346 & 0.089 & 59.3 & 62.6 & 59.6 \\
Qwen2.5-7B Aloe medical\,$+$\,Math$^{\star}$ & 2 & 0.50 & 0.279 & 0.082 & 68.1 & 68.0 & 68.1 \\
Qwen2.5-7B HuatuoGPT\,$+$\,R1-Distill$^{\star}$ & 2 & 0.50 & 0.349 & 0.076 & 68.1 & 68.9 & 68.1 \\
Qwen2.5-7B Math\,$+$\,Coder\,$+$\,Instruct & 3 & 0.33 & 0.443 & 0.042 & 68.0 & 68.6 & 68.1 \\
Mistral-7B MetaMath\,$+$\,Code & 2 & 0.50 & 0.093$^{g}$ & 0.030 & 57.4 & 59.9 & 59.9$^{g}$ \\
Qwen2.5-7B self-SFT, $K\!=\!2$ & 2 & 0.50 & 0.333 & 0.027 & 68.1 & 68.4 & 68.2 \\
Mistral-7B SaulLM legal\,$+$\,BioMistral$^{\star}$ & 2 & 0.50 & 0.322 & 0.025 & 57.3 & 57.8 & 57.4 \\
Qwen2.5-7B Math\,$+$\,EVA roleplay$^{\star}$ & 2 & 0.50 & 0.229 & 0.019 & 68.1 & 68.1 & 68.1 \\
Llama-3.1-8B Swallow ja\,$+$\,Guard$^{\star}$ & 2 & 0.50 & 0.290 & 0.012 & 60.2 & 60.0 & 59.7 \\
Llama-3.1-8B OpenMath2\,$+$\,Hermes-3\,$+$\,Guard & 3 & 0.33 & 0.435 & 0.008 & 60.2 & 60.0 & 59.7 \\
Qwen2.5-7B off-the-shelf $\times6$ & 6 & 0.17 & 0.516 & 0.005 & 68.1 & 68.0 & 68.2 \\
Qwen2.5-7B self-SFT, $K\!=\!8$ & 8 & 0.13 & 0.720 & 0.003 & 68.1 & 68.0 & 68.0 \\
Llama-3.1-8B Tulu-3\,$+$\,Tulu-3-DPO$^{\star}$ & 2 & 0.50 & 0.003$^{g}$ & 0.002 & 59.3 & 64.5 & 64.5$^{g}$ \\
\bottomrule
\end{tabular}
\end{table}

\paragraph{Observed discrimination.} Table~\ref{tab:breadth} and Figure~\ref{fig:predictor}(a--d) cover twenty-two configurations across four families and $K\!\in\![2,8]$. Seven are destructive (perplexity $>100$ or mean $>10$pp below base). Fifteen are harmless (within $1$pp or better), with no intermediate cases. The score exceeds the threshold of $1.9\!\times\!10^{-3}$ for five of the seven destructive merges and for none of the harmless ones. We compute $c\sqrt{\overline{I}/K}$ from each evaluation's scaled task vectors and convention, with $\overline{I}$ the parameter-count-weighted MLP mean interference, detailed in Appendix~\ref{app:breadth_details}.

\paragraph{Rival statistics.} Interference, total power and signal power reach AUC $0.98$ or higher, and interference also sets the repair's variance scale. Mean incoherence $1-\overline{\CR}$ and $\rho$ are anti-predictive.

\paragraph{The gate.} For raw Llama-3.1-8B $K\!=\!3$, gating ($\rho\!=\!0.021$, $\lambda\!=\!0$) raises ungated \prism{} from $59.6$ to Task Arithmetic's $62.0$ (base $60.2$), as Table~\ref{tab:llama_k3} shows. On Mistral-7B $K\!=\!2$ at $\rho\!=\!0.093$, the gate likewise returns Task Arithmetic's $59.9$, $2.5$pp above the base, in Table~\ref{tab:breadth}.

\paragraph{A prospective test.} With the statistic, convention and threshold fixed, twelve of the fourteen predictions made before evaluation were correct across three model families, including the harmless and destructive endpoints of the constructed pair. On the other two merges predicted to be destructive, R1-Distill$+$Coder and Math-PRM$+$Coder, naive averaging falls $18.4$ and $14.4$pp below the base in Table~\ref{tab:breadth}, whereas \prism{} keeps both within $0.5$pp of the base.

\paragraph{A controlled titration.} We continued pretraining two Qwen2.5-1.5B specialists on disjoint Japanese web text and biomedical abstracts. At learning rate $10^{-4}$, eight measured checkpoints follow an approximate square-root law through a harmless endpoint at one billion tokens for each specialist in Figure~\ref{fig:titration}. Extrapolated, the fitted law crosses the threshold near $7.5$ billion tokens and reaches $4.17\!\times\!10^{-3}$, the lower edge of the destructive cluster in Table~\ref{tab:breadth}, near $36$ billion. Continuing at $5\times$ the rate reaches that cluster within a further billion tokens and improves own-domain perplexity (Japanese $16.7\!\to\!12.2$, PubMed $7.2\!\to\!7.0$). As predicted before evaluation, the merge at this endpoint is destructive. Naive averaging loses $20.3$pp against the base, while \prism{} stays within $0.3$pp of it.

\paragraph{The recipe.}\label{sec:recipe} Reported \prism{} results above evaluate the denoising operator, including its $\rho<\tau_g$ bypass gate. Our recipe uses naive averaging below the fixed score threshold and Eq.~\ref{eq:soft_thresh}'s gated shrinkage above. All fifteen harmless configurations in Table~\ref{tab:breadth} scored below the threshold and therefore keep the plain average, which scores $62.6$ against the base's $59.3$ on Llama-3.1-8B R1-Distill with Tulu-3. The twenty-eight public pairs without a code specialist that we screened across three families all scored below half the threshold, as Appendix~\ref{app:breadth_details} details.
\section{Conclusion}
\label{sec:conclusion}

We have shown that the variance of task vectors across specialists both predicts which merges collapse and calibrates their repair. Across twenty-two merges from four model families, the interference score built on this variance exceeded its threshold only for destructive merges, and twelve of the fourteen predictions made before evaluation were correct. \prism{}, which soft-thresholds the average of the task vectors, kept above-threshold destructive merges near the base without data or tuning. Because the score grows with training, model developers can track it to keep their specialists mergeable.

\section*{Ethics Statement}

All authors have read and adhere to the ICLR Code of Ethics. This work merges publicly released checkpoints and specialists we trained from public base models, and it evaluates the merges on public benchmarks. It involves no human subjects and collects no personal data. The continued pretraining uses public Japanese web text from FineWeb-2 and PubMed abstracts, as described in Appendix~\ref{app:breadth_details}. Merging can combine the weaknesses of its constituents along with their capabilities: pruning parameter updates could remove fragile safety alignment or amplify biases that conflict across specialists. Our screen and evaluations target capability benchmarks and do not measure safety behavior, so merged models need separate safety and alignment evaluation before deployment~\citep{lee2022norose}.

\bibliographystyle{iclr2027_conference}
\bibliography{custom}

@inproceedings{hu2022lora,
  title={LoRA: Low-Rank Adaptation of Large Language Models},
  author={Hu, Edward J. and Shen, Yelong and Wallis, Phillip and Allen-Zhu, Zeyuan and Li, Yuanzhi and Wang, Shean and Wang, Lu and Chen, Weizhu},
  booktitle={International Conference on Learning Representations},
  year={2022},
  url={https://openreview.net/forum?id=nZeVKeeFYf9},
  eprint={2106.09685},
  archiveprefix={arXiv}
}

@inproceedings{ilharco2022editing,
  title={Editing Models with Task Arithmetic},
  author={Ilharco, Gabriel and Ribeiro, Marco Tulio and Wortsman, Mitchell and Gururangan, Suchin and Schmidt, Ludwig and Hajishirzi, Hannaneh and Farhadi, Ali},
  booktitle={International Conference on Learning Representations},
  year={2023},
  url={https://openreview.net/forum?id=6t0Kwf8-jrj},
  eprint={2212.04089},
  archiveprefix={arXiv}
}

@inproceedings{yadav2023ties,
  title={TIES-Merging: Resolving Interference When Merging Models},
  author={Yadav, Prateek and Tam, Derek and Choshen, Leshem and Raffel, Colin A. and Bansal, Mohit},
  booktitle={Advances in Neural Information Processing Systems},
  volume={36},
  pages={7093--7115},
  year={2023},
  doi={10.52202/075280-0310},
  url={https://proceedings.neurips.cc/paper_files/paper/2023/hash/1644c9af28ab7916874f6fd6228a9bcf-Abstract-Conference.html},
  eprint={2306.01708},
  archiveprefix={arXiv}
}

@inproceedings{yu2024language,
  title={Language Models Are Super Mario: Absorbing Abilities from Homologous Models as a Free Lunch},
  author={Yu, Le and Yu, Bowen and Yu, Haiyang and Huang, Fei and Li, Yongbin},
  booktitle={Proceedings of the 41st International Conference on Machine Learning},
  volume={235},
  pages={57755--57775},
  year={2024},
  publisher={PMLR},
  url={https://proceedings.mlr.press/v235/yu24p.html},
  eprint={2311.03099},
  archiveprefix={arXiv}
}

@article{chopra2025lewis,
  title={LEWIS (LayEr WIse Sparsity) -- A Training Free Guided Model Merging Approach},
  author={Chopra, Hetarth and Rambhia, Vidhi and Adve, Vikram},
  journal={arXiv preprint arXiv:2503.03874},
  year={2025},
  url={https://arxiv.org/abs/2503.03874},
  eprint={2503.03874},
  archiveprefix={arXiv}
}

@article{yang2023adamerging,
  title={AdaMerging: Adaptive Model Merging for Multi-Task Learning},
  author={Yang, Enneng and Wang, Zhenyi and Shen, Li and Liu, Shiwei and Guo, Guibing and Wang, Xingwei and Tao, Dacheng},
  journal={arXiv preprint arXiv:2310.02575},
  year={2023},
  url={https://arxiv.org/abs/2310.02575},
  eprint={2310.02575},
  archiveprefix={arXiv}
}

@article{huang2024emr,
  title={EMR-Merging: Tuning-Free High-Performance Model Merging},
  author={Huang, Chenyu and Ye, Peng and Chen, Tao and He, Tong and Yue, Xiangyu and Ouyang, Wanli},
  journal={Advances in Neural Information Processing Systems},
  volume={37},
  pages={122741--122769},
  year={2024},
  url={https://arxiv.org/abs/2405.17461},
  eprint={2405.17461},
  archiveprefix={arXiv}
}

@article{deep2024della,
  title={DELLA-Merging: Reducing Interference in Model Merging through Magnitude-Based Sampling},
  author={Deep, Pala Tej and Bhardwaj, Rishabh and Poria, Soujanya},
  journal={arXiv preprint arXiv:2406.11617},
  year={2024},
  url={https://arxiv.org/abs/2406.11617},
  eprint={2406.11617},
  archiveprefix={arXiv}
}

@inproceedings{wortsman2022model,
  title={Model Soups: Averaging Weights of Multiple Fine-Tuned Models Improves Accuracy without Increasing Inference Time},
  author={Wortsman, Mitchell and Ilharco, Gabriel and Gadre, Samir Ya. and Roelofs, Rebecca and Gontijo-Lopes, Raphael and Morcos, Ari S. and Namkoong, Hongseok and Farhadi, Ali and Carmon, Yair and Kornblith, Simon and Schmidt, Ludwig},
  booktitle={Proceedings of the 39th International Conference on Machine Learning},
  volume={162},
  pages={23965--23998},
  year={2022},
  publisher={PMLR},
  url={https://proceedings.mlr.press/v162/wortsman22a.html},
  eprint={2203.05482},
  archiveprefix={arXiv}
}

@article{grattafiori2024llama,
  title={The Llama 3 Herd of Models},
  author={Grattafiori, Aaron and Dubey, Abhimanyu and Jauhri, Abhinav and Pandey, Abhinav and Kadian, Abhishek and Al-Dahle, Ahmad and Letman, Aiesha and Mathur, Akhil and Schelten, Alan and Vaughan, Alex and others},
  journal={arXiv preprint arXiv:2407.21783},
  year={2024},
  url={https://arxiv.org/abs/2407.21783},
  eprint={2407.21783},
  archiveprefix={arXiv}
}

@article{yang2024qwen2,
  title={Qwen2.5-Math Technical Report: Toward Mathematical Expert Model via Self-Improvement},
  author={Yang, An and Zhang, Beichen and Hui, Binyuan and Gao, Bofei and Yu, Bowen and Li, Chengpeng and Liu, Dayiheng and Tu, Jianhong and Zhou, Jingren and Lin, Junyang and others},
  journal={arXiv preprint arXiv:2409.12122},
  year={2024},
  url={https://arxiv.org/abs/2409.12122},
  eprint={2409.12122},
  archiveprefix={arXiv}
}

@article{hui2024qwen2,
  title={Qwen2.5-Coder Technical Report},
  author={Hui, Binyuan and Yang, Jian and Cui, Zeyu and Yang, Jiaxi and Liu, Dayiheng and Zhang, Lei and Liu, Tianyu and Zhang, Jiajun and Yu, Bowen and Lu, Keming and others},
  journal={arXiv preprint arXiv:2409.12186},
  year={2024},
  url={https://arxiv.org/abs/2409.12186},
  eprint={2409.12186},
  archiveprefix={arXiv}
}

@inproceedings{gu2025neuronmerge,
  title={NeuronMerge: Merging Models via Functional Neuron Groups},
  author={Gu, Wangyun and Gao, Qianghua and Zhang, Li-Xin and Shen, Xu and Ye, Jieping},
  booktitle={Findings of the Association for Computational Linguistics: ACL 2025},
  pages={9015--9037},
  year={2025},
  doi={10.18653/v1/2025.findings-acl.471},
  url={https://aclanthology.org/2025.findings-acl.471/}
}

@article{donoho1994ideal,
  title={Ideal Spatial Adaptation by Wavelet Shrinkage},
  author={Donoho, David L. and Johnstone, Iain M.},
  journal={Biometrika},
  volume={81},
  number={3},
  pages={425--455},
  year={1994},
  doi={10.1093/biomet/81.3.425},
  url={https://doi.org/10.1093/biomet/81.3.425}
}

@article{merity2016pointer,
  title={Pointer Sentinel Mixture Models},
  author={Merity, Stephen and Xiong, Caiming and Bradbury, James and Socher, Richard},
  journal={arXiv preprint arXiv:1609.07843},
  year={2016},
  url={https://arxiv.org/abs/1609.07843},
  eprint={1609.07843},
  archiveprefix={arXiv}
}

@inproceedings{matena2022merging,
  title={Merging Models with Fisher-Weighted Averaging},
  author={Matena, Michael S. and Raffel, Colin A.},
  booktitle={Advances in Neural Information Processing Systems},
  volume={35},
  pages={17703--17716},
  year={2022},
  doi={10.52202/068431-1287},
  url={https://doi.org/10.52202/068431-1287}
}

@article{chen2021evaluating,
  title={Evaluating Large Language Models Trained on Code},
  author={Chen, Mark and Tworek, Jerry and Jun, Heewoo and Yuan, Qiming and others},
  journal={arXiv preprint arXiv:2107.03374},
  year={2021}
}

@article{austin2021program,
  title={Program Synthesis with Large Language Models},
  author={Austin, Jacob and Odena, Augustus and Nye, Maxwell and Bosma, Maarten and others},
  journal={arXiv preprint arXiv:2108.07732},
  year={2021}
}

@inproceedings{fang2025see,
  title={To See a World in a Spark of Neuron: Disentangling Multi-Task Interference for Training-Free Model Merging},
  author={Fang, Zitao and Du, Guodong and Yu, Shuyang and Guo, Yifei and Zhang, Yiwei and Cao, Yiyao and Li, Jing and Tang, Ho-Kin and Goh, Sim Kuan},
  booktitle={Proceedings of the 2025 Conference on Empirical Methods in Natural Language Processing},
  pages={15720--15740},
  year={2025},
  doi={10.18653/v1/2025.emnlp-main.793},
  url={https://aclanthology.org/2025.emnlp-main.793/}
}

@inproceedings{ilharco2022patching,
  title={Patching Open-Vocabulary Models by Interpolating Weights},
  author={Ilharco, Gabriel and Wortsman, Mitchell and Gadre, Samir Yitzhak and Song, Shuran and Hajishirzi, Hannaneh and Kornblith, Simon and Farhadi, Ali and Schmidt, Ludwig},
  booktitle={Advances in Neural Information Processing Systems},
  volume={35},
  pages={29262--29277},
  year={2022},
  doi={10.52202/068431-2122},
  url={https://proceedings.neurips.cc/paper_files/paper/2022/hash/bc6cddcd5d325e1c0f826066c1ad0215-Abstract-Conference.html}
}

@inproceedings{fu2025training,
  title={Training-Free LLM Merging for Multi-Task Learning},
  author={Fu, Zichuan and Wu, Xian and Wang, Yejing and Wang, Wanyu and Ye, Shanshan and Yin, Hongzhi and Chang, Yi and Zheng, Yefeng and Zhao, Xiangyu},
  booktitle={Proceedings of the 63rd Annual Meeting of the Association for Computational Linguistics (Volume 1: Long Papers)},
  pages={33111--33124},
  year={2025},
  doi={10.18653/v1/2025.acl-long.1588},
  url={https://aclanthology.org/2025.acl-long.1588/}
}

@inproceedings{liu2026globa,
  title={GLOBA: Rethinking Parameter Conflicts in Model Merging},
  author={Liu, Zehao and Li, Kun and Zhou, Wei},
  booktitle={Proceedings of the AAAI Conference on Artificial Intelligence},
  volume={40},
  pages={23954--23962},
  year={2026},
  doi={10.1609/aaai.v40i28.39572},
  url={https://ojs.aaai.org/index.php/AAAI/article/view/39572}
}

@inproceedings{wang2024localizing,
  title={Localizing Task Information for Improved Model Merging and Compression},
  author={Wang, Ke and Dimitriadis, Nikolaos and Ortiz-Jimenez, Guillermo and Fleuret, François and Frossard, Pascal},
  booktitle={Proceedings of the 41st International Conference on Machine Learning},
  volume={235},
  pages={50268--50287},
  year={2024},
  publisher={PMLR},
  url={https://proceedings.mlr.press/v235/wang24k.html},
  eprint={2405.07813},
  archiveprefix={arXiv}
}

@article{donoho1995noising,
  title={De-noising by soft-thresholding},
  author={Donoho, David L},
  journal={IEEE transactions on information theory},
  volume={41},
  number={3},
  pages={613--627},
  year={1995},
  publisher={IEEE}
}

@inproceedings{xu2024training,
  title={Training-Free Pretrained Model Merging},
  author={Xu, Zhengqi and Yuan, Ke and Wang, Huiqiong and Wang, Yong and Song, Mingli and Song, Jie},
  booktitle={2024 IEEE/CVF Conference on Computer Vision and Pattern Recognition (CVPR)},
  pages={5915--5925},
  year={2024},
  doi={10.1109/CVPR52733.2024.00565},
  url={https://openaccess.thecvf.com/content/CVPR2024/html/Xu_Training-Free_Pretrained_Model_Merging_CVPR_2024_paper.html},
  eprint={2403.01753},
  archiveprefix={arXiv}
}

@inproceedings{sun2025task,
  title={Task Arithmetic in Trust Region: A Training-Free Model Merging Approach to Navigate Knowledge Conflicts},
  author={Sun, Wenju and Li, Qingyong and Wang, Wen and Geng, Yangliao and Li, Boyang},
  booktitle={Proceedings of the 33rd ACM International Conference on Multimedia},
  pages={5178--5187},
  year={2025},
  doi={10.1145/3746027.3755789},
  url={https://doi.org/10.1145/3746027.3755789}
}

@inproceedings{geva2022transformer,
  title={Transformer Feed-Forward Layers Build Predictions by Promoting Concepts in the Vocabulary Space},
  author={Geva, Mor and Caciularu, Avi and Wang, Kevin and Goldberg, Yoav},
  booktitle={Proceedings of the 2022 Conference on Empirical Methods in Natural Language Processing},
  pages={30--45},
  year={2022},
  doi={10.18653/v1/2022.emnlp-main.3},
  url={https://aclanthology.org/2022.emnlp-main.3/}
}

@inproceedings{dai2022knowledge,
  title={Knowledge Neurons in Pretrained Transformers},
  author={Dai, Damai and Dong, Li and Hao, Yaru and Sui, Zhifang and Chang, Baobao and Wei, Furu},
  booktitle={Proceedings of the 60th Annual Meeting of the Association for Computational Linguistics (Volume 1: Long Papers)},
  pages={8493--8502},
  year={2022},
  doi={10.18653/v1/2022.acl-long.581},
  url={https://aclanthology.org/2022.acl-long.581/}
}

@inproceedings{gargiulo2025task,
  title={Task Singular Vectors: Reducing Task Interference in Model Merging},
  author={Gargiulo, Antonio Andrea and Crisostomi, Donato and Bucarelli, Maria Sofia and Scardapane, Simone and Silvestri, Fabrizio and Rodolà, Emanuele},
  booktitle={2025 IEEE/CVF Conference on Computer Vision and Pattern Recognition (CVPR)},
  pages={18695--18705},
  year={2025},
  doi={10.1109/CVPR52734.2025.01742},
  url={https://openaccess.thecvf.com/content/CVPR2025/html/Gargiulo_Task_Singular_Vectors_Reducing_Task_Interference_in_Model_Merging_CVPR_2025_paper.html},
  eprint={2412.00081},
  archiveprefix={arXiv}
}

@inproceedings{goddard2024arcee,
  title={Arcee's MergeKit: A Toolkit for Merging Large Language Models},
  author={Goddard, Charles and Siriwardhana, Shamane and Ehghaghi, Malikeh and Meyers, Luke and Karpukhin, Vladimir and Benedict, Brian and McQuade, Mark and Solawetz, Jacob},
  booktitle={Proceedings of the 2024 Conference on Empirical Methods in Natural Language Processing: Industry Track},
  pages={477--485},
  year={2024},
  doi={10.18653/v1/2024.emnlp-industry.36},
  url={https://aclanthology.org/2024.emnlp-industry.36/}
}

@inproceedings{liu2024evalplus,
  title={Is Your Code Generated by {C}hat{GPT} Really Correct? Rigorous Evaluation of Large Language Models for Code Generation},
  author={Liu, Jiawei and Xia, Chunqiu Steven and Wang, Yuyao and Zhang, Lingming},
  booktitle={Advances in Neural Information Processing Systems},
  volume={36},
  pages={21558--21572},
  year={2023},
  doi={10.52202/075280-0943},
  url={https://proceedings.neurips.cc/paper_files/paper/2023/hash/43e9d647ccd3e4b7b5baab53f0368686-Abstract.html},
  eprint={2305.01210},
  archiveprefix={arXiv}
}

@inproceedings{cheng2025whoever,
  title={Whoever Started the Interference Should End It: Guiding Data-Free Model Merging via Task Vectors},
  author={Cheng, Runxi and Xiong, Feng and Wei, Yongxian and Zhu, Wanyun and Yuan, Chun},
  booktitle={Proceedings of the 42nd International Conference on Machine Learning},
  volume={267},
  pages={10121--10143},
  year={2025},
  publisher={PMLR},
  url={https://proceedings.mlr.press/v267/cheng25h.html},
  eprint={2503.08099},
  archiveprefix={arXiv}
}

@inproceedings{marczak2025iso,
  title={No Task Left Behind: Isotropic Model Merging with Common and Task-Specific Subspaces},
  author={Marczak, Daniel and Magistri, Simone and Cygert, Sebastian and Twardowski, Bartłomiej and Bagdanov, Andrew D. and van de Weijer, Joost},
  booktitle={Proceedings of the 42nd International Conference on Machine Learning},
  volume={267},
  pages={43177--43199},
  year={2025},
  publisher={PMLR},
  url={https://proceedings.mlr.press/v267/marczak25a.html},
  eprint={2502.04959},
  archiveprefix={arXiv}
}

@inproceedings{lu2024twin,
  title={Twin-Merging: Dynamic Integration of Modular Expertise in Model Merging},
  author={Lu, Zhenyi and Fan, Chenghao and Wei, Wei and Qu, Xiaoye and Chen, Dangyang and Cheng, Yu},
  booktitle={Advances in Neural Information Processing Systems},
  volume={37},
  pages={78905--78935},
  year={2024},
  doi={10.52202/079017-2505},
  url={https://proceedings.neurips.cc/paper_files/paper/2024/hash/8fcd17eb91bae20d9826786d7d6be799-Abstract-Conference.html},
  eprint={2406.15479},
  archiveprefix={arXiv}
}

@article{park2026pact,
  title={PACT: Preserving Anchored Cores in Task-Vectors for Model Merging},
  author={Shi, Ningyuan and Zhou, Zhipeng and Wang, Hao and Miao, Chunyan and Zhao, Peilin},
  journal={arXiv preprint arXiv:2606.18627},
  year={2026},
  url={https://arxiv.org/abs/2606.18627},
  eprint={2606.18627},
  archiveprefix={arXiv}
}

@inproceedings{tang2025mergebench,
  title={MergeBench: A Benchmark for Merging Domain-Specialized LLMs},
  author={He, Yifei and Zeng, Siqi and Hu, Yuzheng and Yang, Rui and Zhang, Tong and Zhao, Han},
  booktitle={Advances in Neural Information Processing Systems},
  volume={38},
  year={2025},
  doi={10.52202/085713-3374},
  url={https://proceedings.neurips.cc/paper_files/paper/2025/hash/91f7f71cb04699f387dc863da42a1fe3-Abstract-Datasets_and_Benchmarks_Track.html},
  eprint={2505.10833},
  archiveprefix={arXiv}
}

@article{ortizjimenez2023tangent,
  title={Task arithmetic in the tangent space: Improved editing of pre-trained models},
  author={Ortiz-Jimenez, Guillermo and Favero, Alessandro and Frossard, Pascal},
  journal={Advances in Neural Information Processing Systems},
  volume={36},
  year={2023}
}

@article{yadav2024matters,
  title={What matters for model merging at scale?},
  author={Yadav, Prateek and Vu, Tu and Lai, Jonathan and Chronopoulou, Alexandra and Faruqui, Manaal and Bansal, Mohit and Munkhdalai, Tsendsuren},
  journal={arXiv preprint arXiv:2410.03617},
  year={2024}
}

@inproceedings{jang2024modelstock,
  title={Model Stock: All We Need Is Just a Few Fine-Tuned Models},
  author={Jang, Dong-Hwan and Yun, Sangdoo and Han, Dongyoon},
  booktitle={Computer Vision -- ECCV 2024},
  series={Lecture Notes in Computer Science},
  volume={15102},
  pages={207--223},
  year={2024},
  publisher={Springer Nature Switzerland},
  doi={10.1007/978-3-031-72784-9_12},
  url={https://link.springer.com/chapter/10.1007/978-3-031-72784-9_12}
}

@inproceedings{shoemake1985animating,
  title={Animating Rotation with Quaternion Curves},
  author={Shoemake, Ken},
  booktitle={Proceedings of the 12th Annual Conference on Computer Graphics and Interactive Techniques},
  pages={245--254},
  year={1985},
  doi={10.1145/325334.325242},
  url={https://doi.org/10.1145/325334.325242}
}

@article{yang2024qwen25,
  title={Qwen2.5 Technical Report},
  author={Yang, An and Yang, Baosong and Zhang, Beichen and Hui, Binyuan and Zheng, Bo and Yu, Bowen and Li, Chengyuan and Liu, Dayiheng and Huang, Fei and others},
  journal={arXiv preprint arXiv:2412.15115},
  year={2024},
  url={https://arxiv.org/abs/2412.15115},
  eprint={2412.15115},
  archiveprefix={arXiv}
}

@article{huang2026rain,
  title={RAIN-Merging: A Gradient-Free Method to Enhance Instruction Following in Large Reasoning Models with Preserved Thinking Format},
  author={Huang, Zhehao and Liu, Yuhang and Lin, Baijiong and Lou, Yixin and He, Zhengbao and Tian, Hanling and Li, Tao and Huang, Xiaolin},
  journal={arXiv preprint arXiv:2602.22538},
  year={2026},
  url={https://arxiv.org/abs/2602.22538},
  eprint={2602.22538},
  archiveprefix={arXiv}
}

@inproceedings{gururangan2020dont,
  title={Don't Stop Pretraining: Adapt Language Models to Domains and Tasks},
  author={Gururangan, Suchin and Marasovi{\'c}, Ana and Swayamdipta, Swabha and Lo, Kyle and Beltagy, Iz and Downey, Doug and Smith, Noah A.},
  booktitle={Proceedings of the 58th Annual Meeting of the Association for Computational Linguistics},
  pages={8342--8360},
  year={2020},
  publisher={Association for Computational Linguistics},
  doi={10.18653/v1/2020.acl-main.740},
  url={https://aclanthology.org/2020.acl-main.740/}
}

@misc{lee2022norose,
  title = {There is no rose without a thorn: Finding weaknesses on BlenderBot 2.0 in terms of Model, Data and User-Centric Approach},
  author = {Jungseob Lee and Midan Shim and Suhyune Son and Chanjun Park and Yujin Kim and Heuiseok Lim},
  year = {2022},
  note = {arXiv preprint arXiv:2201.03239},
  eprint = {2201.03239},
  archivePrefix = {arXiv},
  url = {https://arxiv.org/abs/2201.03239},
}

@article{cobbe2021training,
  title={Training Verifiers to Solve Math Word Problems},
  author={Cobbe, Karl and Kosaraju, Vineet and Bavarian, Mohammad and Chen, Mark and Jun, Heewoo and Kaiser, Lukasz and Plappert, Matthias and Tworek, Jerry and Hilton, Jacob and Nakano, Reiichiro and Hesse, Christopher and Schulman, John},
  journal={arXiv preprint arXiv:2110.14168},
  year={2021},
  url={https://arxiv.org/abs/2110.14168},
  eprint={2110.14168},
  archiveprefix={arXiv}
}

@article{clark2018think,
  title={Think you have Solved Question Answering? {T}ry {ARC}, the {AI2} Reasoning Challenge},
  author={Clark, Peter and Cowhey, Isaac and Etzioni, Oren and Khot, Tushar and Sabharwal, Ashish and Schoenick, Carissa and Tafjord, Oyvind},
  journal={arXiv preprint arXiv:1803.05457},
  year={2018},
  url={https://arxiv.org/abs/1803.05457},
  eprint={1803.05457},
  archiveprefix={arXiv}
}

@inproceedings{hendrycks2021measuring,
  title={Measuring Massive Multitask Language Understanding},
  author={Hendrycks, Dan and Burns, Collin and Basart, Steven and Zou, Andy and Mazeika, Mantas and Song, Dawn and Steinhardt, Jacob},
  booktitle={International Conference on Learning Representations},
  year={2021},
  url={https://openreview.net/forum?id=d7KBjmI3GmQ}
}

@inproceedings{lin2022truthfulqa,
  title={{TruthfulQA}: Measuring How Models Mimic Human Falsehoods},
  author={Lin, Stephanie and Hilton, Jacob and Evans, Owain},
  booktitle={Proceedings of the 60th Annual Meeting of the Association for Computational Linguistics (Volume 1: Long Papers)},
  year={2022},
  doi={10.18653/v1/2022.acl-long.229},
  url={https://doi.org/10.18653/v1/2022.acl-long.229}
}

@inproceedings{zellers2019hellaswag,
  title={{HellaSwag}: Can a Machine Really Finish Your Sentence?},
  author={Zellers, Rowan and Holtzman, Ari and Bisk, Yonatan and Farhadi, Ali and Choi, Yejin},
  booktitle={Proceedings of the 57th Annual Meeting of the Association for Computational Linguistics},
  year={2019},
  doi={10.18653/v1/P19-1472},
  url={https://doi.org/10.18653/v1/P19-1472}
}

@inproceedings{sakaguchi2020winogrande,
  title={{WinoGrande}: An Adversarial {W}inograd Schema Challenge at Scale},
  author={Sakaguchi, Keisuke and Le Bras, Ronan and Bhagavatula, Chandra and Choi, Yejin},
  booktitle={Proceedings of the AAAI Conference on Artificial Intelligence},
  year={2020},
  doi={10.1609/aaai.v34i05.6399},
  url={https://doi.org/10.1609/aaai.v34i05.6399}
}

@misc{gao2024language,
  title={The Language Model Evaluation Harness},
  author={Gao, Leo and Tow, Jonathan and Abbasi, Baber and Biderman, Stella and Black, Sid and DiPofi, Anthony and Foster, Charles and Golding, Laurence and Hsu, Jeffrey and Le Noac'h, Alain and Li, Haonan and McDonell, Kyle and Muennighoff, Niklas and Ociepa, Chris and Phang, Jason and Reynolds, Laria and Schoelkopf, Hailey and Skowron, Aviya and Sutawika, Lintang and Tang, Eric and Thite, Anish and Wang, Ben and Wang, Kevin and Zou, Andy},
  howpublished={Zenodo},
  note={Version v0.4.3},
  year={2024},
  doi={10.5281/zenodo.12608602},
  url={https://zenodo.org/records/12608602}
}

@article{lightman2023lets,
  title={Let's Verify Step by Step},
  author={Lightman, Hunter and Kosaraju, Vineet and Burda, Yura and Edwards, Harri and Baker, Bowen and Lee, Teddy and Leike, Jan and Schulman, John and Sutskever, Ilya and Cobbe, Karl},
  journal={arXiv preprint arXiv:2305.20050},
  year={2023}
}

@article{cao2026qwen3coder,
  title={{Qwen3-Coder-Next} Technical Report},
  author={Cao, Ruisheng and Chen, Mouxiang and Chen, Jiawei and Cui, Zeyu and Feng, Yunlong and Hui, Binyuan and Jing, Yuheng and Li, Kaixin and Li, Mingze and Lin, Junyang and others},
  journal={arXiv preprint arXiv:2603.00729},
  year={2026},
  url={https://arxiv.org/abs/2603.00729},
  eprint={2603.00729},
  archiveprefix={arXiv}
}

@article{cohere2025command,
  title={Command {A}: An Enterprise-Ready Large Language Model},
  author={{Team Cohere} and Aakanksha and Ahmadian, Arash and Ahmed, Marwan and Alammar, Jay and Alizadeh, Milad and others},
  journal={arXiv preprint arXiv:2504.00698},
  year={2025},
  url={https://arxiv.org/abs/2504.00698},
  eprint={2504.00698},
  archiveprefix={arXiv}
}

@inproceedings{wang2026scaling,
  title={Model Merging Scaling Laws in Large Language Models},
  author={Wang, Yuanyi and Gu, Yanggan and Zhang, Yiming and Zhou, Qi and Yan, Zhaoyi and Xie, Congkai and Wang, Xinyao and Yuan, Jianbo and Yang, Hongxia},
  booktitle={Proceedings of the 43rd International Conference on Machine Learning},
  year={2026},
  url={https://openreview.net/forum?id=VkTfx8FJoq},
  eprint={2509.24244},
  archiveprefix={arXiv}
}

@article{hitit2026systematic,
  title={A Systematic Study of In-the-Wild Model Merging for Large Language Models},
  author={Hitit, O{\u{g}}uz Ka{\u{g}}an and Girrbach, Leander and Akata, Zeynep},
  journal={Transactions on Machine Learning Research},
  year={2026},
  url={https://openreview.net/forum?id=6zSIyrqS7J},
  eprint={2511.21437},
  archiveprefix={arXiv}
}

@inproceedings{krogh1994neural,
  title={Neural Network Ensembles, Cross Validation, and Active Learning},
  author={Krogh, Anders and Vedelsby, Jesper},
  booktitle={Advances in Neural Information Processing Systems},
  volume={7},
  year={1994},
  url={https://proceedings.neurips.cc/paper_files/paper/1994/hash/b8c37e33defde51cf91e1e03e51657da-Abstract.html}
}

\appendix

\section{Full PRISM Algorithm}
\label{app:algorithm}

Algorithm~\ref{alg:prism} gives the complete \prism{}-Denoise procedure summarized in Section~\ref{sec:method} and illustrated in Figure~\ref{fig:teaser}.

\begin{algorithm}[ht]
\caption{\prism{}-Denoise: Interference-Calibrated Soft Thresholding}
\label{alg:prism}
\begin{algorithmic}[1]
\REQUIRE Base $\theta_{\text{base}}$, specialists $\{\theta_k\}_{k=1}^K$, threshold scale $\kappa$, coefficient $c$, gate $\tau_g$
\ENSURE Merged model $\theta_{\text{merged}}$
\STATE \textbf{// Step 1: Task Vectors and Norm Scaling}
\STATE $\tau_k \leftarrow \theta_k-\theta_{\text{base}}$ for all $k$; restrict to merged MLP coordinates
\STATE $s_k \leftarrow \min_j\|\tau_j\|/\|\tau_k\|$ for all $k$ \hfill (or $s_k=1$ for raw merges)
\STATE \textbf{// Step 2: Aggregation, Layer Interference, and Cancellation}
\FOR{each MLP layer $l$, containing $n_l$ merged coordinates}
    \STATE $\bar{\delta}_l \leftarrow \frac{1}{K}\sum_k s_k\tau_{k,l}$
    \STATE $\sigma_l \leftarrow \sqrt{\overline{I}_l}=\sqrt{\overline{P}_l-\overline{S}_l}$ \hfill (empirical interference scale)
    \STATE $\overline{C}_l \leftarrow \text{mean}_i\,\min\!\big(\tfrac{2}{K}\min(\text{Pos}(i),\text{Neg}(i))^2,\,I(i)\big)$
    \STATE $\lambda_l \leftarrow \kappa\sigma_l\sqrt{2\log n_l}$
\ENDFOR
\STATE \textbf{// Step 3: Bypass Thresholding When Cancellation Is Low}
\STATE $\rho \leftarrow \big(\sum_l n_l\overline{C}_l\big)/\big(\sum_l n_l\overline{I}_l+\epsilon_g\big)$
\IF{$\rho<\tau_g$}
    \STATE $\lambda_l\leftarrow0$ for all $l$ \hfill (Task Arithmetic with the same $s_k$ and $c$)
\ENDIF
\STATE \textbf{// Step 4: Soft Thresholding with a Keep-Rate Floor}
\STATE $\tilde{\delta}\leftarrow0$; $p_{\min}\leftarrow0.1\%$
\FOR{each MLP layer $l$}
    \FOR{each projection matrix $m$ in layer $l$, containing $n_{l,m}$ coordinates}
        \STATE $\tilde{\delta}_{l,m}\leftarrow\operatorname{sgn}(\bar{\delta}_{l,m})\max(0,|\bar{\delta}_{l,m}|-\lambda_l)$ \hfill (elementwise)
        \IF{$\|\tilde{\delta}_{l,m}\|_0/n_{l,m}<p_{\min}$}
            \STATE $k_{l,m}\leftarrow\max(\lfloor p_{\min}n_{l,m}\rfloor,1)$
            \STATE $q_{l,m}\leftarrow$ the $k_{l,m}$-th largest magnitude in $|\bar{\delta}_{l,m}|$
            \STATE $\tilde{\delta}_{l,m}\leftarrow\operatorname{sgn}(\bar{\delta}_{l,m})\max(0,|\bar{\delta}_{l,m}|-q_{l,m})$
        \ENDIF
    \ENDFOR
\ENDFOR
\STATE \textbf{return} $\theta_{\text{base}}+c\tilde{\delta}$
\end{algorithmic}
\end{algorithm}

The floor is checked separately for each MLP projection matrix. Numerical safeguards clamp the threshold variance at $10^{-12}$ and the logarithm's element count at $2$; the gate denominator uses $\epsilon_g=10^{-12}$. Norm scaling uses a $10^{-10}$ denominator stabilizer and returns unit scales when all task vectors vanish. Layer coherence also uses $10^{-10}$ in its denominator; $\overline{I}_l=\overline{P}_l-\overline{S}_l$ is computed directly, with negative roundoff clipped to zero.

\section{Denoise-Once-Then-Amplify: Full Argument}
\label{app:amplify_proof}

Proposition~\ref{prop:amplify} in Section~\ref{sec:injected} combines an exact scaling identity with a working noise model. Equation~\ref{eq:interference_var} defines $I=P-S$ as the empirical variance across scaled task vectors.

For the working model, write $s_k\tau_k(i)=\mu(i)+\eta_k(i)$, with zero-mean disturbances that are uncorrelated across specialists and have a common variance $v_l$ at a coordinate of layer $l$. Then $\operatorname{Var}(K^{-1}\sum_k\eta_k(i))=v_l/K$. Applying $c$ after this average gives standard deviation $c\sqrt{v_l/K}$ and variance $c^2v_l/K$. Replacing the unknown $v_l$ by the measured $\overline{I}_l$ gives the plug-in screening scale $c\sqrt{\overline{I}_l/K}$ and its squared value $c^2\overline{I}_l/K$. In the implementation the mean already divides by $K$: $c=1$ writes the full average, and our default $c=1/K$ shrinks it further.

For any fixed update, multiplication by $c$ scales its squared norm by $c^2$, and \prism{} thresholds the average before this multiplication. In the coefficient sweep of Section~\ref{sec:c_sweep}, baselines deteriorate at $c\!\ge\!0.7$, with the worst reaching perplexity $961$ at $c=0.7$, while \prism{} retains low perplexity across the tested sweep.

\section{Mechanism Analysis}
\label{app:mechanism}

\begin{figure}[ht]
\centering
\includegraphics[width=\linewidth]{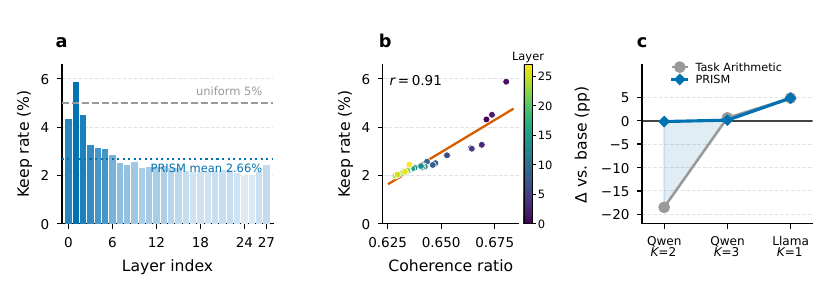}
\caption{Mechanism on Qwen2.5-7B Math+Coder ($K\!=\!2$, instruction-tuned specialists, $\kappa\!=\!0.5$). \textbf{(a)} Layer-wise retention, bars shaded by coherence. \textbf{(b)} Retention against coherence, colored by depth. \textbf{(c)} Distance from the base model across settings ordered by interference.}
\label{fig:mechanism}
\end{figure}

Figure~\ref{fig:mechanism} examines the Qwen2.5-7B Math+Coder merge at $\kappa\!=\!0.5$. Here the calibrated threshold sets the keep rate above the $0.1\%$ floor: the mean is $2.66\%$ with instruction-tuned specialists in panel~a and $2.3\%$ with base specialists in Table~\ref{tab:ablation}. At the default $\kappa\!=\!1$, the floor binds in every layer and retention is approximately flat. At $\kappa\!=\!0.5$, early layers with $\overline{\CR}_l\!\approx\!0.68$ keep $4$--$6\%$ of coordinates, while middle layers with $\overline{\CR}_l\!\approx\!0.63$ keep about $2\%$. Retention $p_l$ and coherence have Pearson $r\!=\!0.91$ in panel b.

Across the settings in panel c, the advantage over Task Arithmetic is largest on Math+Coder, whose sign agreement is $58\%$ and disagreement $42\%$, and vanishes on the Llama-3.1-8B single-specialist merge, where $\CR\!=\!1$. For fixed layer power and dimension, $\CR\!\to\!1$ implies $\sigma_l\!\to\!0$ up to the numerical stabilizer, and a zero threshold gives Task Arithmetic with the same scaling.

Coherence mixes two sources of disagreement, which the split of Section~\ref{sec:cancel} separates. Ignoring the stabilizer, mutual conflict ($\tau_1\!=\!+10,\tau_2\!=\!-10$) gives $\CR\!=\!0$, while a task-exclusive update ($\tau_1\!=\!+10,\tau_2\!=\!0$) gives $\CR\!=\!0.5$. Shrinking the latter can remove useful specialist signal, and the $\rho$-gate of Section~\ref{sec:gate} returns the plain average when dispersion dominates. Interference-based scoring exceeds activation-norm scoring by $1.3$pp without calibration data, as Appendix~\ref{app:lewis_ctp} details.

\section{Extended Results}
\label{app:extended}

\subsection{Calibration Granularity}
\label{app:granularity}

Table~\ref{tab:granularity} compares global, row-wise and layer-wise estimates of the threshold scale in Eq.~\ref{eq:soft_thresh}, together with a sign-agreement substitute. Their six-task means lie within evaluation uncertainty, and their perplexities range from $9.0$ to $9.1$. Global and layer-wise estimates stay close at larger coefficients, with $67.8$ against $67.9$ at $c\!=\!0.7$ and $65.0$ against $64.8$ at $c\!=\!1.0$. The same holds on the six- and eight-specialist merges of Section~\ref{sec:breadth}, with $67.9$ against $68.2$ and $68.2$ against $68.0$. We retain the layer-wise default because it lets the threshold vary across depth.

\begin{table}[ht]
\caption{Calibration-granularity variants on Qwen2.5-7B Math+Coder ($K\!=\!2$, $c\!=\!0.5$). The variants use interference or sign agreement to calibrate a soft threshold.}
\label{tab:granularity}
\centering
\small
\begin{tabular}{lcc}
\toprule
\textbf{Variant} & \textbf{Mean} & \textbf{PPL}$\downarrow$ \\
\midrule
Global-$\lambda$ (single $\sigma$) & 68.1 & 9.0 \\
SA-calibrated $\sigma$ (sign agreement) & 67.8 & 9.0 \\
Row-wise $\sigma$ (empirical Bayes) & 67.8 & 9.1 \\
\rowcolor{gray!15} \prism{} (layer-wise $\CR$, default) & 67.8 & 9.0 \\
\bottomrule
\end{tabular}
\end{table}

\subsection{Threshold Scale and Scope}

Table~\ref{tab:ablation} sweeps the threshold scale $\kappa$ and the merged scope on Qwen2.5-7B $K\!=\!2$. The $0.1\%$ floor binds in every layer for $\kappa\!\geq\!1$, so $\kappa\!=\!1$, $2$ and $3$ give identical merged weights and share one row.

With the floor disabled at $\kappa\!=\!1$, the calibrated threshold keeps $0.01\%$ of coordinates and writes $1.4\%$ of the average's norm. The six-task mean is $68.6$ at perplexity $8.9$, within evaluation uncertainty of the floored operator. On HumanEval, HumanEval$+$, MBPP and MBPP$+$, the \textsc{EvalPlus} scores fall from the floored $65.9/59.1/71.2/61.1$ to $63.4/57.3/70.6/60.8$, so the floor adds $0.3$ to $2.5$pp on each coding metric.

\begin{table}[ht]
\caption{Threshold-scale and scope ablations of \prism{} on Qwen2.5-7B $K\!=\!2$. \emph{floor}: the $0.1\%$ keep-rate floor binds in every layer. \textbf{Bold} = best in each block.}
\label{tab:ablation}
\centering
\small
\setlength{\tabcolsep}{3pt}
\begin{tabular}{llccccccc@{\hskip 4pt}c}
\toprule
\textbf{Ablation} & \textbf{Variant} & \textbf{PPL}$\downarrow$ & \textbf{GSM} & \textbf{ARC} & \textbf{MMLU} & \textbf{TQA} & \textbf{HS} & \textbf{WG} & \textbf{Mean} \\
\midrule
\multirow{3}{*}{\shortstack[l]{Threshold\\scale $\kappa$}}
& $\kappa\!=\!0.25$ (keep $22.6\%$) & 11.6 & 81.2 & \textbf{56.6} & 72.0 & 53.9 & 66.6 & 74.2 & 67.4 \\
& $\kappa\!=\!0.5$ (keep $2.3\%$) & 9.5 & 79.0 & 53.2 & 74.1 & 54.2 & \textbf{67.8} & 75.4 & 67.3 \\
& $\kappa\!\in\![1,3]$ (floor, default) & \textbf{9.0} & \textbf{81.4} & 51.4 & \textbf{74.3} & \textbf{55.2} & 67.4 & \textbf{77.0} & \textbf{67.8} \\
\midrule
\multirow{2}{*}{Scope}
& MLP $+$ Attention & \textbf{9.0} & \textbf{83.0} & 51.2 & 74.2 & \textbf{55.5} & \textbf{67.6} & 76.0 & \textbf{67.9} \\
& MLP-only (default) & \textbf{9.0} & 81.4 & \textbf{51.4} & \textbf{74.3} & 55.2 & 67.4 & \textbf{77.0} & 67.8 \\
\bottomrule
\end{tabular}
\end{table}

\subsection{Interference Distribution}
\label{app:interference_dist}

We examine the distribution on which the threshold of Eq.~\ref{eq:soft_thresh} acts. The layer statistic $\sigma_l^2=\overline{I}_l$ averages interference across $n_l\!\sim\!10^8$ coordinates in the Qwen2.5-7B Math$+$Coder merge. Across its 28 MLP layers, the sampled standardized aggregate has excess kurtosis $0.44$ and skewness near zero, with mild heavy-tail deviations beyond $|z|\!>\!3$ on a Q-Q plot. Layer coherence correlates negatively with mean interference ($r\!=\!-0.41$).

Under the common-component, independent-disturbance model, the empirical variance with divisor $K$ has expectation $(K{-}1)/K$ times the disturbance variance, including for $K\!\in\!\{2,3\}$. A constant rescaling of $\sigma_l$ can be offset algebraically by $\kappa$ before the floor is applied. Using $\sigma_l$ rather than $\sigma_l/\sqrt{K}$ changes the threshold by $\sqrt{K}$, at most about $2.8$ over the studied range through $K=8$. The $\kappa$ sweep in Table~\ref{tab:ablation} spans a factor of $12$, wider than any of these convention changes, and the six-task mean stays between $67.3$ and $67.8$ across it.

At $K\!=\!2$, an element-level $\sigma(i)$ rests on a single squared difference. Multiplying this unstable local scale by the universal factor $\sqrt{2\log n}\!\approx\!6.2$ risks excessive shrinkage. The row-wise estimate already matches the layer-wise six-task mean in Table~\ref{tab:granularity}.

\subsection{Specialist Checkpoints and Evaluation Uncertainty}
\label{app:multi_seed}\label{app:instruct_panel}

Table~\ref{tab:stderr} compares instruction-tuned and base specialist checkpoints, with benchmark-level standard errors from binomial proportion variance at the shared task-specific sample sizes. Both specialist pairs are merged into Qwen2.5-7B with task vectors taken against it, and the instruction-tuned pair consists of Math-7B-Instruct and Coder-7B-Instruct. \prism{} leads the listed merges with both pairs, and Task Arithmetic scores $6.5$ points higher with instruction-tuned specialists.

Individual benchmarks carry standard errors of $1.7$ to $2.2$pp, with MMLU at $0.4$pp. Propagating the six task-specific errors gives at most $0.8$pp for a six-task mean and roughly $1.1$pp for a difference between two means. 
\begin{table}[ht]
\caption{Benchmark scores with standard errors for Qwen2.5-7B Math$+$Coder merges of instruction-tuned and base specialists; the base-specialist merges are those of Table~\ref{tab:k2_results}. The final column propagates the six benchmark errors into the standard error of the mean.}
\label{tab:stderr}
\centering
\small
\setlength{\tabcolsep}{2.5pt}
\begin{tabular}{lccccccc}
\toprule
\textbf{Method} & \textbf{GSM8K} & \textbf{ARC} & \textbf{MMLU} & \textbf{TQA} & \textbf{HS} & \textbf{WG} & \textbf{Mean} \\
\midrule
\multicolumn{8}{l}{\emph{Instruction-tuned specialists}} \\
Task Arith. & $47.8\pm2.2$ & $48.4\pm2.2$ & $59.5\pm0.4$ & $53.0\pm2.0$ & $57.8\pm2.2$ & $69.8\pm2.1$ & $56.0\pm0.8$ \\
TIES & $69.4\pm2.1$ & $51.4\pm2.2$ & $65.4\pm0.4$ & $52.8\pm2.0$ & $62.6\pm2.2$ & $70.0\pm2.1$ & $61.9\pm0.8$ \\
DELLA & $70.4\pm2.0$ & $52.2\pm2.2$ & $65.0\pm0.4$ & $51.7\pm2.0$ & $62.2\pm2.2$ & $70.6\pm2.0$ & $62.0\pm0.8$ \\
LEWIS (wt.-sp.) & $77.4\pm1.9$ & $54.6\pm2.2$ & $70.6\pm0.4$ & $50.7\pm2.0$ & $66.2\pm2.1$ & $72.0\pm2.0$ & $65.2\pm0.8$ \\
\rowcolor{gray!15} \prism{} & $81.2\pm1.7$ & $51.4\pm2.2$ & $74.3\pm0.4$ & $55.0\pm1.9$ & $67.8\pm2.1$ & $75.8\pm1.9$ & $67.6\pm0.7$ \\
\midrule
\multicolumn{8}{l}{\emph{Base-model specialists}} \\
Task Arith. & $22.0\pm1.9$ & $47.0\pm2.2$ & $55.7\pm0.4$ & $50.0\pm2.0$ & $55.4\pm2.2$ & $66.6\pm2.1$ & $49.5\pm0.8$ \\
TIES & $67.2\pm2.1$ & $53.4\pm2.2$ & $64.3\pm0.4$ & $50.2\pm2.0$ & $60.6\pm2.2$ & $71.0\pm2.0$ & $61.1\pm0.8$ \\
DELLA & $65.2\pm2.1$ & $51.6\pm2.2$ & $64.2\pm0.4$ & $50.3\pm2.0$ & $59.8\pm2.2$ & $70.4\pm2.0$ & $60.3\pm0.8$ \\
LEWIS (act.) & $43.0\pm2.2$ & $44.2\pm2.2$ & $69.1\pm0.4$ & $50.7\pm2.0$ & $62.8\pm2.2$ & $66.6\pm2.1$ & $56.1\pm0.8$ \\
\rowcolor{gray!15} \prism{} & $81.4\pm1.7$ & $51.4\pm2.2$ & $74.3\pm0.4$ & $55.2\pm1.9$ & $67.4\pm2.1$ & $77.0\pm1.9$ & $67.8\pm0.7$ \\
\rowcolor{gray!8} Base model & $82.8\pm1.7$ & $51.4\pm2.2$ & $74.3\pm0.4$ & $55.6\pm1.9$ & $67.0\pm2.1$ & $77.0\pm1.9$ & $68.0\pm0.7$ \\
\bottomrule
\end{tabular}
\end{table}

\subsection{PRISM-Denoise against PRISM-Prune}
\label{app:prune_comparison}

Table~\ref{tab:denoise_vs_prune} compares \prism{}-Denoise with \prism{}-Prune, which averages the scaled task vectors and keeps the largest-magnitude coordinates of each layer at a rate of $5\%$ scaled by the layer's coherence relative to the whole model. With two specialists, Denoise leads on the six-task mean by $3.7$pp at 7B and $6.9$pp at 1.5B, and the variants tie at $K\!=\!3$, where interference is lower. At $K\!=\!1$ Denoise equals Task Arithmetic because $\lambda_l\!=\!0$, while Prune keeps $5\%$ of the task vector and trails by $3.2$pp.

\begin{table}[ht]
\caption{\prism{}-Denoise against \prism{}-Prune across settings (\%). \textbf{Bold} marks the better variant, with ties sharing the mark.}
\label{tab:denoise_vs_prune}
\centering
\resizebox{\textwidth}{!}{
\setlength{\tabcolsep}{2.2pt}
\begin{tabular}{llccccccc cccc}
\toprule
 & & \multicolumn{7}{c}{\textbf{Harness}} & \multicolumn{4}{c}{\textbf{Generation}} \\
\cmidrule(lr){3-9} \cmidrule(lr){10-13}
\textbf{Setting} & \textbf{Variant} & \textbf{GSM8K} & \textbf{ARC} & \textbf{MMLU} & \textbf{TruthQA} & \textbf{HellaSwag} & \textbf{WinoGr.} & \textbf{Mean} & \textbf{GSM8K} & \textbf{MATH} & \textbf{HumanEval} & \textbf{MBPP} \\
\midrule
\rowcolor{gray!15} \cellcolor{white}\multirow{2}{*}{Qwen 7B $K\!=\!2$}
& Denoise & \textbf{81.4} & 51.4 & \textbf{74.3} & \textbf{55.2} & \textbf{67.4} & \textbf{77.0} & \textbf{67.8} & \textbf{82.4} & \textbf{30.2} & \textbf{51.2} & \textbf{62.8} \\
& Prune & 73.8 & \textbf{53.6} & 70.2 & 50.8 & 65.2 & 71.0 & 64.1 & 73.8 & 15.6 & 31.1 & 39.6 \\
\midrule
\rowcolor{gray!15} \cellcolor{white}\multirow{2}{*}{Qwen 1.5B $K\!=\!2$}
& Denoise & \textbf{59.6} & \textbf{42.6} & \textbf{61.8} & 45.5 & \textbf{59.0} & \textbf{66.4} & \textbf{55.8} & \textbf{59.0} & \textbf{26.6} & \textbf{35.4} & \textbf{43.0} \\
& Prune & 35.4 & 40.0 & 53.4 & \textbf{46.2} & 53.8 & 64.8 & 48.9 & 35.6 & 13.8 & 21.3 & 37.4 \\
\midrule
\rowcolor{gray!15} \cellcolor{white}\multirow{2}{*}{Qwen 7B $K\!=\!3$}
& Denoise & \textbf{83.4} & 51.6 & \textbf{74.3} & 55.6 & \textbf{66.8} & \textbf{76.8} & \textbf{68.1} & \textbf{83.6} & 37.2 & 57.3 & \textbf{64.2} \\
& Prune & 83.0 & \textbf{52.0} & \textbf{74.3} & \textbf{55.8} & 66.4 & 76.6 & 68.0 & 82.6 & \textbf{38.0} & \textbf{57.9} & 63.6 \\
\midrule
\rowcolor{gray!15} \cellcolor{white}\multirow{2}{*}{Llama $K\!=\!1$}
& Denoise & \textbf{71.8} & 53.6 & \textbf{68.9} & \textbf{51.3} & 70.6 & \textbf{74.0} & \textbf{65.0} & \textbf{73.4} & \textbf{28.6} & \textbf{53.7} & \textbf{52.4} \\
& Prune & 59.0 & \textbf{54.0} & 67.1 & 46.7 & \textbf{71.2} & 73.0 & 61.8 & 58.8 & 25.0 & 40.2 & 52.2 \\
\bottomrule
\end{tabular}
}
\end{table}

\subsection{Model Scale}
\label{app:scale}

Table~\ref{tab:scale} evaluates merges of the Qwen2.5 Instruct and Coder-Instruct models at 3B and 14B. The interference scores of both merges lie far below the threshold of Table~\ref{tab:breadth}, so the recipe of Section~\ref{sec:recipe} keeps their plain averages, which gain $0.8$ and $1.0$pp over the base.

\begin{table}[ht]
\caption{Instruct$+$Coder-Instruct merges of Qwen2.5 at 3B and 14B ($K\!=\!2$, $c\!=\!0.5$, six-task mean, \%). Both score far below the threshold of Table~\ref{tab:breadth}, so the recipe returns the plain average (TA).}
\label{tab:scale}
\centering
\small
\begin{tabular}{lccc}
\toprule
\textbf{Model} & $c\sqrt{\overline{I}/K}$ $(\times10^{-3})$ & \textbf{Base} & \textbf{TA} \\
\midrule
Qwen2.5-3B & 0.076 & 60.7 & 61.5 \\
Qwen2.5-14B & 0.067 & 71.9 & 72.9 \\
\bottomrule
\end{tabular}
\end{table}

\subsection{LEWIS Variants and the Scoring Ablation}
\label{app:lewis_act}\label{app:lewis_ctp}

In Table~\ref{tab:lewis_act}, activation-based LEWIS trails the weight-sparsity variant by $8.8$pp on the Qwen2.5-7B Math$+$Coder merge of Table~\ref{tab:k2_results}. The gap narrows to $0.8$pp on the Llama-3.1-8B single-specialist merge and closes on the Qwen2.5-7B three-specialist merge, where interference is lower. Table~\ref{tab:lewis_ctp} compares activation-norm and interference-based threshold scales on the same Math$+$Coder merge while retaining soft thresholding. Denoising with the activation-based scale scores $10.4$pp above activation-based pruning, and \prism{}'s interference-based scale adds a further $1.3$pp without calibration data.

\begin{table}[ht]
\caption{Activation-based and weight-sparsity LEWIS on three merges (six-task mean, \%). \textbf{Bold} marks the better variant in each merge.}
\label{tab:lewis_act}
\centering
\small
\begin{tabular}{llc}
\toprule
\textbf{Setting} & \textbf{Method} & \textbf{Mean (\%)} \\
\midrule
\multirow{3}{*}{Qwen $K\!=\!2$}
& LEWIS (act.) & 56.1 \\
& LEWIS (wt.-sp.) & \textbf{64.9} \\
& Base Model & 68.0 \\
\midrule
\multirow{3}{*}{Llama $K\!=\!1$}
& LEWIS (act.) & 60.6 \\
& LEWIS (wt.-sp.) & \textbf{61.4} \\
& Base Model & 60.0 \\
\midrule
\multirow{3}{*}{Qwen $K\!=\!3$}
& LEWIS (act.) & \textbf{68.1} \\
& LEWIS (wt.-sp.) & \textbf{68.1} \\
& Base Model & 68.0 \\
\bottomrule
\end{tabular}
\end{table}

\begin{table}[ht]
\caption{Scoring ablation on Qwen2.5-7B Math$+$Coder base specialists ($K\!=\!2$, six-task mean).}
\label{tab:lewis_ctp}
\centering
\small
\begin{tabular}{lcc}
\toprule
\textbf{Scoring Method} & \textbf{Mean (\%)} & \textbf{PPL}$\downarrow$ \\
\midrule
\rowcolor{gray!15} \prism{}-Denoise (interference $\sigma$) & \textbf{67.8} & \textbf{9.0} \\
LEWIS-CTP (activation $\sigma$ + denoise) & 66.5 & 11.9 \\
LEWIS-Act (activation + prune) & 56.1 & 24.2 \\
\bottomrule
\end{tabular}
\end{table}

\subsection{Data-Dependent Baselines}
\label{app:fisher_soup}

Table~\ref{tab:datadep} lists the data-dependent baselines on Qwen $K\!=\!2$. AdaMerging learns layer-wise coefficients by minimizing prediction entropy on 128 unlabeled C4 texts with 100 steps of simultaneous perturbation stochastic approximation (SPSA). It reaches parity with \prism{} only at its tuned coefficient and falls to $51.6$ at the shared $c\!=\!0.5$. Greedy Soup, which adds a specialist only when it lowers perplexity on 50 WikiText-2 texts, rejects both specialists and returns the base model. Fisher-weighted averaging, which estimates a diagonal Fisher on 100 WikiText-2 texts, falls to $47.5$.

\begin{table}[ht]
\caption{Data-dependent baselines on Qwen2.5-7B Math+Coder ($K\!=\!2$, six-task mean). \prism{} uses no data.}
\label{tab:datadep}
\centering
\small
\begin{tabular}{llc}
\toprule
\textbf{Method} & \textbf{Data used} & \textbf{Mean (\%)} \\
\midrule
Base model, unmerged & none & 68.1 \\
AdaMerging (tuned, $c\!=\!0.1$) & 128 C4 texts & 67.9 \\
AdaMerging ($c\!=\!0.5$) & 128 C4 texts & 51.6 \\
Greedy Soup & 50 WikiText-2 texts & 68.1 \\
Fisher-weighted averaging & 100 WikiText-2 texts & 47.5 \\
\rowcolor{gray!15} \prism{} ($c\!=\!0.5$) & none & 67.8 \\
\bottomrule
\end{tabular}
\end{table}

\subsection{Baseline Fairness Protocol}
\label{app:fairness}

Table~\ref{tab:fairness_grid} lists the settings and sweeps used for every baseline across our experiments. \prism{} uses $\kappa\!=\!1.0$ without setting-specific tuning. The main task-vector comparisons fix $c\!=\!1/K$ after averaging, and Section~\ref{sec:c_sweep} reports the coefficient sweep. The GSM8K and MATH-500 generation evaluations fix the TIES trim at $k\!=\!0.2$ and use weight-sparsity LEWIS, which needs no calibration data, whereas activation-based LEWIS computes its layer importance from 128 WikiText-2 texts.

\begin{table}[ht]
\caption{Baseline settings and sweeps across all experiments. The LEWIS data entry applies to the activation-based variant.}
\label{tab:fairness_grid}
\centering
\small
\begin{tabular}{lll}
\toprule
\textbf{Method} & \textbf{Hyperparameters} & \textbf{Data used} \\
\midrule
Task Arithmetic & $c\in\{0.1,0.3,0.5,0.7,1.0\}$ & none \\
TIES & $c$ (as above); trim $k\in\{0.1,0.2\}$ & none \\
DELLA & $c$; $\epsilon\!=\!0.3$; trim $k\!=\!0.2$ & none \\
LEWIS & $c$; keep rate $\in\{0.05,0.1\}$ & 128 WikiText-2 texts \\
Fisher & $c$; diagonal Fisher & 100 WikiText-2 texts \\
Greedy Soup & inclusion by perplexity & 50 WikiText-2 texts \\
AdaMerging & $c\in\{0.1,0.2,0.3,0.5\}$; 100 SPSA steps; layer-wise coeffs & 128 C4 texts \\
Iso-C & $\alpha\in\{0.10,0.25,0.50\}$ & none \\
DARE & drop rate $p\in\{0.5,0.9\}$ at $c\!=\!0.5$; $c\!=\!1.0$ at $p\!=\!0.9$ & none \\
WUDI-Merging & scope $\in$ \{MLP, all linear\}; lr $\in\{10^{-5},10^{-6}\}$; 300 iters & none \\
\rowcolor{gray!15} \prism{} (ours) & none (fixed $\kappa\!=\!1.0$) & none \\
\bottomrule
\end{tabular}
\end{table}

\subsection{Why Baselines Fail on Qwen \texorpdfstring{$K\!=\!2$}{K=2}}
\label{app:why_baselines}

The Math$+$Coder task vectors are large and conflict in sign on about $42\%$ of elements. This weakens sign consensus for TIES and DELLA and motivates examining the size of the update written by each method. At $c\!=\!0.5$, DARE collapses at both tested drop rates: PPL $7.9\!\times\!10^{5}$ with a six-task mean of $30.0$ at $p\!=\!0.9$, and PPL $128$ with $39.2$ at $p\!=\!0.5$. Its $1/(1-p)$ rescaling enlarges the surviving coordinates, and at $p\!=\!0.9$ its update reaches $2.85\times$ the norm of the plain average in Table~\ref{tab:extended_baselines}.

\textbf{Iso-C and WUDI-Merging}, the two 2025 data-free methods, were evaluated over the grids in Table~\ref{tab:fairness_grid}. Iso-C at its default coefficient $c\!=\!1/K\!=\!0.5$ collapses with PPL $3.5\!\times\!10^{4}$ and a six-task mean of $29.3$. At $c\!=\!0.25$ it gives PPL $36.4$ and a mean of $44.3$; at $c\!=\!0.1$ it reaches PPL $11.6$ and a mean of $66.6$, $1.2$pp below untuned \prism{}. Its scores are $78.4/54.6/71.4/55.0/64.8/75.2$ on GSM8K, ARC, MMLU, TruthfulQA, HellaSwag and WinoGrande. WUDI-Merging diverged in all four evaluated configurations, with six-task means of $29.6$ to $30.3$. MLP scope at learning rate $10^{-5}$ gave PPL $2.1\!\times\!10^{7}$, and all linear layers at $10^{-5}$ gave $1.6\!\times\!10^{6}$. On the instruction-tuned versions of these specialists, all linear layers gave $1.3\!\times\!10^{6}$ at $10^{-5}$ and $2.2\!\times\!10^{6}$ at $10^{-6}$.

\textbf{Community recipes} are our evaluations on the same specialist pair, using each recipe's own rule without our external coefficient or norm scaling. SLERP at $t\!=\!0.5$ \citep{shoemake1985animating}, as used in the mergekit recipe \citep{goddard2024arcee}, is numerically close to the full unscaled average because the shared base dominates each tensor. It writes $1.27\times$ the norm-scaled mean and collapses at PPL $1.0\!\times\!10^{6}$. Model Stock \citep{jang2024modelstock} derives $t\!\approx\!0.41$ from the observed mean task-vector cosine of $0.26$, giving $2.4\times$ shrinkage, and still loses $14.8$pp at perplexity $35$. \prism{}, by comparison, writes $2.2\%$ of the average's norm in Table~\ref{tab:extended_baselines}. Model Stock's angle-derived rule does not directly calibrate the absolute update magnitude, consistent with the limitations of direction statistics measured in Section~\ref{sec:breadth}.

\subsection{Scaling-Coefficient Sweep}
\label{app:csweep_harness}

Table~\ref{tab:csweep} gives the numbers behind Figure~\ref{fig:c_sweep}: the six-task harness mean and the four generation benchmarks at every coefficient. The unmerged base has a harness mean of $68.1$ and scores $83.8$, $38.2$, $56.7$ and $63.8$ on GSM8K, MATH-500, HumanEval and MBPP. At $c\!=\!0.7$, baseline perplexities are $961$, $100$, $116$ and $105$ for Task Arithmetic, TIES, DELLA and LEWIS, and at $c\!=\!1.0$ they are $4.6\!\times\!10^{5}$, $1.1\!\times\!10^{4}$, $1.6\!\times\!10^{4}$ and $5.9\!\times\!10^{3}$.

\begin{table}[ht]
\caption{Scaling-coefficient sweep on Qwen2.5-7B Math$+$Coder: six-task harness mean and generation accuracy (\%). LEWIS: activation-based harness mean, weight-sparsity generation. \textbf{Bold} = best merged in each block.}
\label{tab:csweep}\label{tab:csweep_harness}
\centering
\small
\setlength{\tabcolsep}{4pt}
\begin{tabular}{llccccc}
\toprule
& & \textbf{Harness} & \multicolumn{4}{c}{\textbf{Generation}} \\
\cmidrule(lr){3-3}\cmidrule(lr){4-7}
\textbf{$c$} & \textbf{Method} & \textbf{Mean} & \textbf{GSM8K} & \textbf{MATH-500} & \textbf{HumanEval} & \textbf{MBPP} \\
\midrule
\multirow{5}{*}{$0.3$}
& Task Arith. & 63.4 & 69.2 & 29.0 & 33.5 & 48.8 \\
& TIES & 66.1 & 78.4 & 30.2 & 42.1 & 53.8 \\
& DELLA & 65.8 & 78.2 & 27.6 & 42.7 & 53.4 \\
& LEWIS & 67.6 & 80.8 & 22.6 & 41.5 & 55.0 \\
\rowcolor{gray!15} & \prism{}-Denoise & \textbf{68.0} & \textbf{83.0} & \textbf{32.6} & \textbf{51.2} & \textbf{64.8} \\
\midrule
\multirow{5}{*}{$0.5$}
& Task Arith. & 49.5 & 23.0 & 5.8 & 13.4 & 27.4 \\
& TIES & 61.1 & 65.8 & 24.6 & 17.7 & 43.0 \\
& DELLA & 60.3 & 64.0 & 22.0 & 17.7 & 42.4 \\
& LEWIS & 56.1 & 74.2 & 15.4 & 31.7 & 41.4 \\
\rowcolor{gray!15} & \prism{}-Denoise & \textbf{67.8} & \textbf{82.4} & \textbf{30.2} & \textbf{51.2} & \textbf{62.8} \\
\midrule
\multirow{5}{*}{$0.7$}
& Task Arith. & 35.1 & 0.2 & 0.0 & 1.2 & 1.0 \\
& TIES & 49.3 & 32.2 & 10.0 & 6.7 & 13.6 \\
& DELLA & 48.3 & 33.2 & 9.6 & 11.6 & 11.0 \\
& LEWIS & 49.4 & 62.8 & 6.4 & 23.2 & 25.8 \\
\rowcolor{gray!15} & \prism{}-Denoise & \textbf{67.9} & \textbf{81.8} & \textbf{27.0} & \textbf{47.0} & \textbf{61.2} \\
\midrule
\multirow{5}{*}{$1.0$}
& Task Arith. & 28.8 & 0.0 & 0.0 & 0.0 & 0.0 \\
& TIES & 30.4 & 0.0 & 0.0 & 0.0 & 0.0 \\
& DELLA & 30.9 & 0.0 & 0.0 & 0.0 & 0.0 \\
& LEWIS & 38.7 & 17.4 & 0.4 & 1.8 & 1.2 \\
\rowcolor{gray!15} & \prism{}-Denoise & \textbf{64.8} & \textbf{65.8} & \textbf{12.6} & \textbf{39.0} & \textbf{57.2} \\
\bottomrule
\end{tabular}
\end{table}

\subsection{Low-Rank Specialists}
\label{app:lora}

LoRA represents an update as a low-rank product \citep{hu2022lora}. A trained adapter can be materialized as $\tau_k=(\alpha/r)\mathbf{B}_k\mathbf{A}_k$ before merging in full weight space. The experiment in Table~\ref{tab:lora_merge} instead simulates low-rank updates by retaining the top-$r$ singular components of each full Math+Coder task vector. Across ranks from $8$ to $128$, \prism{} stays within $0.6$pp of its full-weight six-task mean of $67.8$ and outperforms Task Arithmetic at every rank on all three measures.

\begin{table}[ht]
\caption{Low-rank SVD simulation on Qwen2.5-7B Math+Coder ($K\!=\!2$) across projection ranks: six-task harness mean, GSM8K generation accuracy and perplexity for \prism{}-Denoise and Task Arithmetic (TA), with the merge's coherence ratio.}
\label{tab:lora_merge}
\centering
\small
\setlength{\tabcolsep}{5pt}
\begin{tabular}{lc cc cc cc}
\toprule
& & \multicolumn{2}{c}{\textbf{Mean}} & \multicolumn{2}{c}{\textbf{GSM8K (gen.)}} & \multicolumn{2}{c}{\textbf{PPL}$\downarrow$} \\
\cmidrule(lr){3-4}\cmidrule(lr){5-6}\cmidrule(lr){7-8}
\textbf{Rank} & $\CR$ & \prism{} & TA & \prism{} & TA & \prism{} & TA \\
\midrule
$r\!=\!8$ & 0.529 & \textbf{67.7} & 65.1 & \textbf{81.8} & 78.2 & \textbf{10.1} & 14.2 \\
$r\!=\!16$ & 0.524 & \textbf{67.9} & 65.2 & \textbf{83.8} & 78.4 & \textbf{9.9} & 14.5 \\
$r\!=\!32$ & 0.520 & \textbf{68.4} & 64.8 & \textbf{84.2} & 77.4 & \textbf{9.8} & 14.9 \\
$r\!=\!64$ & 0.519 & \textbf{68.0} & 64.2 & \textbf{83.4} & 73.6 & \textbf{9.5} & 16.0 \\
$r\!=\!128$ & 0.521 & \textbf{68.3} & 63.2 & \textbf{83.6} & 72.6 & \textbf{9.2} & 17.3 \\
Full fine-tuning & 0.630 & \textbf{67.8} & 49.5 & \textbf{82.4} & 23.0 & \textbf{9.0} & 46.7 \\
\bottomrule
\end{tabular}
\end{table}

\section{Coherence Ratio and Sign Agreement}
\label{app:cr_vs_sa}

For $K\!=\!2$ with equal scaling and nonzero task-vector power, neglecting the stabilizer gives $\CR(i)=(\tau_1(i)+\tau_2(i))^2/\big(2(\tau_1(i)^2+\tau_2(i)^2)\big)$. When both magnitudes are equal and nonzero, this equals binary sign agreement; unequal magnitudes give intermediate values. At the layer level, coherence and sign agreement are highly correlated ($r\!=\!0.998$ on Qwen2.5-7B $K\!=\!2$), an empirical finding rather than a consequence of averaging alone over $n_l\!\sim\!10^8$ coordinates. Coherence relates the observed interference to power through $\sigma_l^2=\overline{I}_l=\overline{P}_l-\overline{S}_l\simeq\overline{P}_l(1-\overline{\CR}_l)$, where the final expression neglects the stabilizer. This supplies the chosen threshold scale; a sign-agreement fraction alone does not determine the magnitude of interference.

\section{Conflict and Exclusivity}
\label{app:rescue}

Section~\ref{sec:analysis} notes that low coherence mixes two phenomena. We make this operational with an element-wise exclusivity score,
\begin{equation}
e(i) = \frac{\max_k |s_k \tau_k(i)|}{\sum_k |s_k \tau_k(i)| + \epsilon},
\end{equation}
For nonzero coordinates and negligible $\epsilon$, $e(i)$ lies in $[1/K,1]$: values near $1$ indicate concentration in one specialist, while values near $1/K$ indicate similar magnitudes across specialists. Equal magnitudes imply symmetric conflict only within the low-$\CR$ subset. When every task-vector entry is zero, $e(i)=0$. Partitioning the low-$\CR$ mass of Qwen2.5-7B Math+Coder by this score shows that thresholded elements are predominantly conflict elements, while many exclusive elements retain a nonzero consensus and survive.

\paragraph{Why the threshold uses the total interference.} Calibrating with cancellation alone, $\sigma_l^2\!\leftarrow\!\overline{C}_l$, gives $67.5$ at $c\!=\!0.5$, $67.0$ at $c\!=\!0.7$ and $60.8$ at $c\!=\!1.0$ on Qwen2.5-7B $K\!=\!2$. At the last coefficient, \prism{} scores $64.8$. Since $\overline{C}_l\le\overline{I}_l$, this substitution lowers the nominal threshold and can retain a larger update. The observed loss at aggressive scaling supports retaining $\sigma_l^2=\overline{I}_l$ for the threshold and using $\rho$ as a binary gate.

\paragraph{Norm scaling with heterogeneous specialists.} The $\min$-norm rule equalizes scaled task-vector norms at the smallest input norm. With strongly heterogeneous specialists, this shrinks the whole update: on the Llama-3.1-8B $K\!=\!3$ merge, the rule keeps every method within $0.5$pp of the base's $60.2$ in Table~\ref{tab:breadth}. Table~\ref{tab:llama_k3} therefore uses unscaled task vectors. The Qwen2.5 Math$+$Coder specialists have comparable norms, with $s_k\!\in\![0.84,1.0]$.

\section{Implementation Details}
\label{app:implementation}

The default matched comparison applies \prism{} and its baselines to the MLP projections (\texttt{gate\_proj}, \texttt{up\_proj}, \texttt{down\_proj}). Attention projections (\texttt{q/k/v/o\_proj}), embeddings, layer norms and the output head retain their base values. Scope ablations and extended recipes state their exceptions. MLP layers can store domain knowledge as key--value memories \citep{geva2022transformer,dai2022knowledge}, which motivates this scope. MLP-only merging stays within $0.1$pp of the attention-inclusive variant's six-task mean in Table~\ref{tab:ablation} with $2.3\times$ fewer merged keys, 84 against 196. The norm-scaling factors on Qwen2.5-7B Math+Coder are $0.84$ and $1.0$.

We load models one at a time, with specialist weights and intermediates on CPU; the peak GPU footprint is a single 7B model, about 15 GB in bfloat16.

\section{Coding Generation}
\label{app:coding}

Table~\ref{tab:coding} reports coding results with the instruction-tuned specialists of Section~\ref{sec:k2_results}. \prism{}-Denoise matches the base model's $63.4$ on MBPP and scores highest on both benchmarks among the merges that use no calibration data.

\begin{table}[ht]
\caption{Coding results with instruction-tuned Qwen2.5-7B Math+Coder specialists ($K\!=\!2$), pass@1 (\%). \textit{\textcolor{gray}{Gray italic}} marks specialist reference scores. LEWIS uses its weight-sparsity variant. \textbf{Bold} best merged, \underline{underline} second best; $^\dagger$needs calibration data.}
\label{tab:coding}
\centering
\small
\begin{tabular}{lcc}
\toprule
\textbf{Method} & \textbf{MBPP} & \textbf{HumanEval} \\
\midrule
Base Model & 63.4 & 57.3 \\
\textit{\textcolor{gray}{Math Specialist}} & \textit{\textcolor{gray}{0.0}} & \textit{\textcolor{gray}{48.8}} \\
\textit{\textcolor{gray}{Coder Specialist}} & \textit{\textcolor{gray}{70.0}} & \textit{\textcolor{gray}{81.7}} \\
\midrule
Task Arith. & 34.8 & 11.6 \\
TIES & 45.1 & 18.3 \\
DELLA & 45.7 & 18.9 \\
LEWIS & 45.7 & 35.4 \\
AdaMerging$^\dagger$ ($c\!=\!0.1$) & \underline{62.8} & \textbf{53.0} \\
\rowcolor{gray!15} \prism{}-Prune & 45.1 & 31.7 \\
\rowcolor{gray!15} \textbf{\prism{}-Denoise} & \textbf{63.4} & \underline{48.8} \\
\bottomrule
\end{tabular}
\end{table}

\section{Coding Evaluation with \texorpdfstring{\textsc{EvalPlus}}{EvalPlus}}
\label{app:evalplus}

We evaluate the same model checkpoints with the EvalPlus framework \citep{liu2024evalplus}, adapting the QwenCoder-2.5 evaluation scaffold in the QwenLM/Qwen3-Coder repository.\footnote{QwenLM, \href{https://github.com/QwenLM/Qwen3-Coder/tree/33bc6aabd7791ad7b32f7e92104f11f2359ba890/qwencoder-eval/instruct/eval_plus}{\texttt{eval\_plus/} scaffold}, commit \texttt{33bc6aa}.} Generation uses vLLM with a fixed ChatML prompt whose assistant turn is pre-seeded with \texttt{```python}, greedy decoding, 2048 new tokens and \texttt{max\_model\_len}=4096. Scoring applies \texttt{evalplus.sanitize} and \texttt{evalplus.evaluate} to the base and augmented (plus) splits. The recipe differs from the \texttt{lm-eval-harness} evaluation in Section~\ref{sec:k2_results} in its prompt, code extraction and test suites. Tables~\ref{tab:extended_baselines} and \ref{tab:evalplus} report the merged models and specialist references under this recipe.

Under this recipe, Task Arithmetic, TIES, DELLA, DARE, LEWIS, SLERP and Model Stock produce degenerate generations, such as runs of zeros, multilingual token noise, empty code blocks or repeated prompt text, and none of them solves more than one problem on any split.

\begin{table}[ht]
\caption{\textsc{EvalPlus} pass@1 (\%) of the unmerged base, \prism{} and the two specialists on Qwen2.5-7B Math+Coder ($K\!=\!2$, base specialists). Table~\ref{tab:extended_baselines} reports the baselines under the same recipe.}
\label{tab:evalplus}
\centering
\small
\setlength{\tabcolsep}{5pt}
\begin{tabular}{lcccc}
\toprule
\textbf{Model} & \textbf{HumanEval} & \textbf{HumanEval+} & \textbf{MBPP} & \textbf{MBPP+} \\
\midrule
Base model, unmerged & 62.8 & 57.9 & 69.8 & 59.5 \\
\rowcolor{gray!15} \prism{}-Denoise & \textbf{65.9} & \textbf{59.1} & \textbf{71.2} & \textbf{61.1} \\
\textit{\textcolor{gray}{Math Specialist}} & \textit{\textcolor{gray}{66.5}} & \textit{\textcolor{gray}{63.4}} & \textit{\textcolor{gray}{60.3}} & \textit{\textcolor{gray}{52.9}} \\
\textit{\textcolor{gray}{Coder Specialist}} & \textit{\textcolor{gray}{64.6}} & \textit{\textcolor{gray}{60.4}} & \textit{\textcolor{gray}{72.0}} & \textit{\textcolor{gray}{60.6}} \\
\bottomrule
\end{tabular}
\end{table}

\section{Breadth Experiments: Details}
\label{app:breadth_details}

\paragraph{Configurations and threshold.} Table~\ref{tab:breadth} reports twenty-two evaluated configurations, seven destructive and fifteen harmless. We placed the threshold of $1.9\!\times\!10^{-3}$ in the gap between the harmless and destructive scores of the nine unstarred configurations, which were measured first. Twelve of the fourteen predictions made before evaluation were correct.

Three settings extend the Qwen2.5-7B Math$+$Coder merge along orthogonal axes: a different base family (Llama-3.1-8B, $K\!=\!3$) in Table~\ref{tab:llama_k3}, an off-the-shelf merge of six community specialists with conflicting objectives (Qwen2.5-7B $K\!=\!6$: Math, Coder, Finance, Thai, Safety, Roleplay), and a matched-recipe self-SFT scaling curve to $K\!=\!8$ in Table~\ref{tab:sft_scaling}.

\paragraph{Screened public pairs.} The screen scored twenty-eight public pairs without a code specialist, fourteen on Qwen2.5-7B, eleven on Llama-3.1-8B and three on Mistral-7B. Their specialists span mathematics and reasoning, medicine, finance, law, long-context modeling, general post-training, safety classification, role-play, and continued pretraining in Japanese, Thai, Southeast Asian and European languages. The scores range from $0.002\!\times\!10^{-3}$ to $0.853\!\times\!10^{-3}$, all below half the threshold, and the largest belongs to Qwen2.5-7B R1-Distill$+$Math-PRM.

\paragraph{Continued-pretraining recipe.} The titration uses full-parameter causal-language-model training from \texttt{Qwen/Qwen2.5-1.5B}. The Japanese run uses the \texttt{train} split of \texttt{HuggingFaceFW/fineweb-2}, configuration \texttt{jpn\_Jpan}; the biomedical run uses titles and abstract content from \texttt{MedRAG/pubmed}, also \texttt{train}. We shuffle the streams with a 10,000-document buffer, insert EOS between documents, and pack 4096-token sequences into an effective batch of 64 sequences. Both phases use fused AdamW ($\beta_1\!=\!0.9$, $\beta_2\!=\!0.999$, $\epsilon\!=\!10^{-8}$), weight decay $0.01$, gradient-norm clipping at $1.0$, bfloat16 computation, and 100 warmup steps followed by a constant learning rate. At $10^{-4}$, the measured steps are $500,1000,\ldots,3500,3815$, totaling $1.00$ billion tokens for each specialist. The higher-rate phase loads each step-3815 checkpoint's weights, restarts the optimizer and step count, and uses $5\!\times\!10^{-4}$. Before shuffling, it skips the first one million Japanese or four million PubMed documents. Its measured steps are $500,1000,\ldots,4000$; step 4000 adds $1.05$ billion tokens to each specialist. At every measured step, the Japanese and biomedical checkpoints are scored as one merge with $K\!=\!2$, $c\!=\!0.5$ and $\min$-norm scaling. The own-domain perplexities in Section~\ref{sec:breadth} use the FineWeb-2 Japanese \texttt{test} split and a PubMed \texttt{train} slice after nine million documents.

\paragraph{The first titration endpoint.} At one billion tokens and learning rate $10^{-4}$, the merge is harmless at a score of $0.699\!\times\!10^{-3}$, with six-task means of $55.4$ for Task Arithmetic, $56.2$ for \prism{} and $56.1$ for the base.

\paragraph{Rival predictors.} Table~\ref{tab:naive_predictors} compares the magnitude statistics $c\sqrt{x/K}$ for $x\in\{\overline{I},\overline{P},\overline{S}\}$, using parameter-count-weighted layer means, with two direction statistics, the mean incoherence $1-\overline{\CR}$ and the cancellation ratio $\rho$. Inverting the definition of $\overline{\CR}_l$ gives total power as $\overline{P}_l=(\overline{I}_l+\epsilon\,\overline{\CR}_l)/(1-\overline{\CR}_l)$, and signal power is $\overline{S}=\overline{P}-\overline{I}$. All three magnitude statistics reach AUC $0.98$ or higher, and total and signal power separate all twenty-two outcomes. Both direction statistics place destructive merges below many harmless ones.

\begin{table}[ht]
\caption{Rival pre-merge statistics on all configurations of Table~\ref{tab:breadth}. AUC treats larger scores as more destructive. Margin is the smallest destructive value divided by the largest harmless one.}
\label{tab:naive_predictors}
\centering
\small
\setlength{\tabcolsep}{6pt}
\begin{tabular}{lccl}
\toprule
\textbf{Statistic} & \textbf{AUC} & \textbf{Margin} & \textbf{Result} \\
\midrule
$c\sqrt{\overline{I}/K}$ (used) & 0.98 & $0.83\times$ & predictive \\
$c\sqrt{\overline{P}/K}$ (total power) & 1.00 & $21.2\times$ & separates \\
$c\sqrt{\overline{S}/K}$ (signal power) & 1.00 & $23.3\times$ & separates \\
$1-\overline{\CR}$ (mean incoherence) & 0.08 & $<\!0.01\times$ & anti-predictive \\
$\rho$ (cancellation ratio) & 0.25 & $0.02\times$ & anti-predictive \\
\bottomrule
\end{tabular}
\end{table}

\begin{table}[ht]
\caption{Llama-3.1-8B $K\!=\!3$ merge of OpenMath2, Hermes-3 and Llama-Guard with unscaled task vectors (six-task mean, \%). The data-free $\rho$-gate detects the dispersion-dominated regime at $\rho\!=\!0.021$ and returns Task Arithmetic. \textbf{Bold} = best.}
\label{tab:llama_k3}
\centering
\small
\begin{tabular}{lc}
\toprule
\textbf{Method} & \textbf{Mean (\%)} \\
\midrule
Base Model & 60.2 \\
\midrule
Task Arithmetic & \textbf{62.0} \\
DELLA & 61.4 \\
TIES & 61.2 \\
LEWIS & 60.3 \\
\rowcolor{gray!15} \prism{}-Denoise (no gate) & 59.6 \\
\rowcolor{gray!15} \prism{}-Denoise ($+\,\rho$-gate) & \textbf{62.0} \\
\bottomrule
\end{tabular}
\end{table}

\paragraph{\texorpdfstring{$K\!=\!6$}{K=6} and self-SFT scaling.} On the six-specialist merge, the base, Task Arithmetic and \prism{} score $68.1$, $68.0$ and $68.2$, consistent with an interference score of $0.005\!\times\!10^{-3}$.

The matched-recipe curve in Table~\ref{tab:sft_scaling} probes expert count under a common training recipe: as $K$ grows from $2$ to $8$, $\rho$ rises monotonically ($0.34\!\to\!0.70$), while every method remains within a point of the base. Neither increasing $K$ nor increasing $\rho$ indicates harm within this curve.

\begin{table}[ht]
\caption{Matched-recipe self-SFT scaling on Qwen2.5-7B with $c\!=\!1/K$ and unscaled task vectors, merging $K\!=\!2$ to $8$ specialists trained from one base under an identical recipe.}
\label{tab:sft_scaling}
\centering
\small
\setlength{\tabcolsep}{5pt}
\begin{tabular}{lccccccc}
\toprule
\textbf{Method} & \textbf{$K\!=\!2$} & \textbf{$K\!=\!3$} & \textbf{$K\!=\!4$} & \textbf{$K\!=\!5$} & \textbf{$K\!=\!6$} & \textbf{$K\!=\!7$} & \textbf{$K\!=\!8$} \\
\midrule
Base Model (ref.) & 68.1 & 68.1 & 68.1 & 68.1 & 68.1 & 68.1 & 68.1 \\
Task Arithmetic & 68.7 & 68.5 & 68.0 & 68.1 & 68.1 & 68.1 & 68.0 \\
TIES & 68.4 & 68.2 & 68.2 & 68.1 & 67.9 & 68.1 & 68.2 \\
\rowcolor{gray!15} \prism{}-Denoise ($+\,\rho$-gate) & 67.9 & 68.1 & 68.1 & 68.1 & 68.0 & 68.0 & 68.1 \\
\midrule
$\rho$ (cancellation ratio) & 0.34 & 0.42 & 0.46 & 0.55 & 0.60 & 0.65 & 0.70 \\
\bottomrule
\end{tabular}
\end{table}

\end{document}